\documentclass[12pt,letterpaper]{article}
\usepackage[a4paper, total={7in, 10in}]{geometry}

\usepackage{graphicx}
\usepackage{helvet}
\usepackage{authblk}
\usepackage{hyperref}
\usepackage{amsmath} 
\usepackage{amssymb} 
\usepackage{orcidlink} 
\usepackage[super,comma,sort&compress]  
   {natbib}
\usepackage[right]{lineno} \linenumbers
\usepackage{booktabs}
\usepackage{multirow}
\usepackage{listings}
\usepackage{color}
\usepackage{xcolor}
\usepackage{xspace}

\colorlet{punct}{black}
\definecolor{background}{HTML}{fafafa}
\definecolor{delim}{RGB}{3, 3, 3}
\colorlet{numb}{black}

\lstdefinelanguage{json}{
    basicstyle=\normalfont\ttfamily,
    showstringspaces=false,
    breaklines=true,
    backgroundcolor=\color{background},
    literate=
     *{0}{{{\color{numb}0}}}{1}
      {1}{{{\color{numb}1}}}{1}
      {2}{{{\color{numb}2}}}{1}
      {3}{{{\color{numb}3}}}{1}
      {4}{{{\color{numb}4}}}{1}
      {5}{{{\color{numb}5}}}{1}
      {6}{{{\color{numb}6}}}{1}
      {7}{{{\color{numb}7}}}{1}
      {8}{{{\color{numb}8}}}{1}
      {9}{{{\color{numb}9}}}{1}
      {:}{{{\color{punct}{:}}}}{1}
      {,}{{{\color{punct}{,}}}}{1}
      {\{}{{{\color{delim}{\{}}}}{1}
      {\}}{{{\color{delim}{\}}}}}{1}
      {[}{{{\color{delim}{[}}}}{1}
      {]}{{{\color{delim}{]}}}}{1},
}

\makeatletter
\renewcommand{\maketitle}{\bgroup\setlength{\parindent}{0pt}
\begin{flushleft}
  \textbf{\@title}
  
  \@author
\end{flushleft}\egroup}
\makeatother

\title{Aligning Large Language Models and Geometric Deep Models for Protein Representation}
\date{}

\author[1]{Dong Shu}
\author[2]{Bingbing Duan}
\author[3]{Kai Guo}
\author[4]{Kaixiong Zhou}
\author[3]{Jiliang Tang}
\author[5]{Mengnan Du*}

\affil[1]{Northwestern University, Computer Science Dept, Evanston, IL, 60201, USA}
\affil[2]{University of Pittsburgh, Biological Sciences Dept, Pittsburgh, PA, 15260, USA}
\affil[3]{Michigan State University, Computer Science Dept, East Lansing, MI, 48824, USA}
\affil[4]{North Carolina State University, Electrical and Computer Engineering Dept, Raleigh, NC, 27695, USA}
\affil[5]{New Jersey Institute of Technology, Data Science Dept, Newark, NJ, 07102, USA}

\affil[*]{Correspondence: mengnan.du@njit.edu}
\affil[]{Lead: dongshu2024@u.northwestern.edu}

\begin{document}

\maketitle

\section*{Summary}

In this study, we explore the alignment of multimodal representations between LLMs and Geometric Deep Models (GDMs) in the protein domain. We comprehensively evaluate three LLMs 
with four protein-specialized GDMs. 
Our work examines alignment factors from both model and protein perspectives, identifying challenges in current alignment methodologies and proposing strategies to improve the alignment process. Experimental results reveal that GDMs incorporating both graph and 3D structural information align better with LLMs, larger LLMs demonstrate improved alignment capabilities, and protein rarity significantly impacts alignment performance. We also find that increasing GDM embedding dimensions, using two-layer projection heads, and fine-tuning LLMs on protein-specific data substantially enhance alignment quality. Lastly, we demonstrate that improved alignment correlates with better downstream performance and reduced hallucination in protein-focused MLLMs. 




\section*{Introduction}

\begin{figure}[t]
    \centering
    \includegraphics[width=0.95\linewidth]{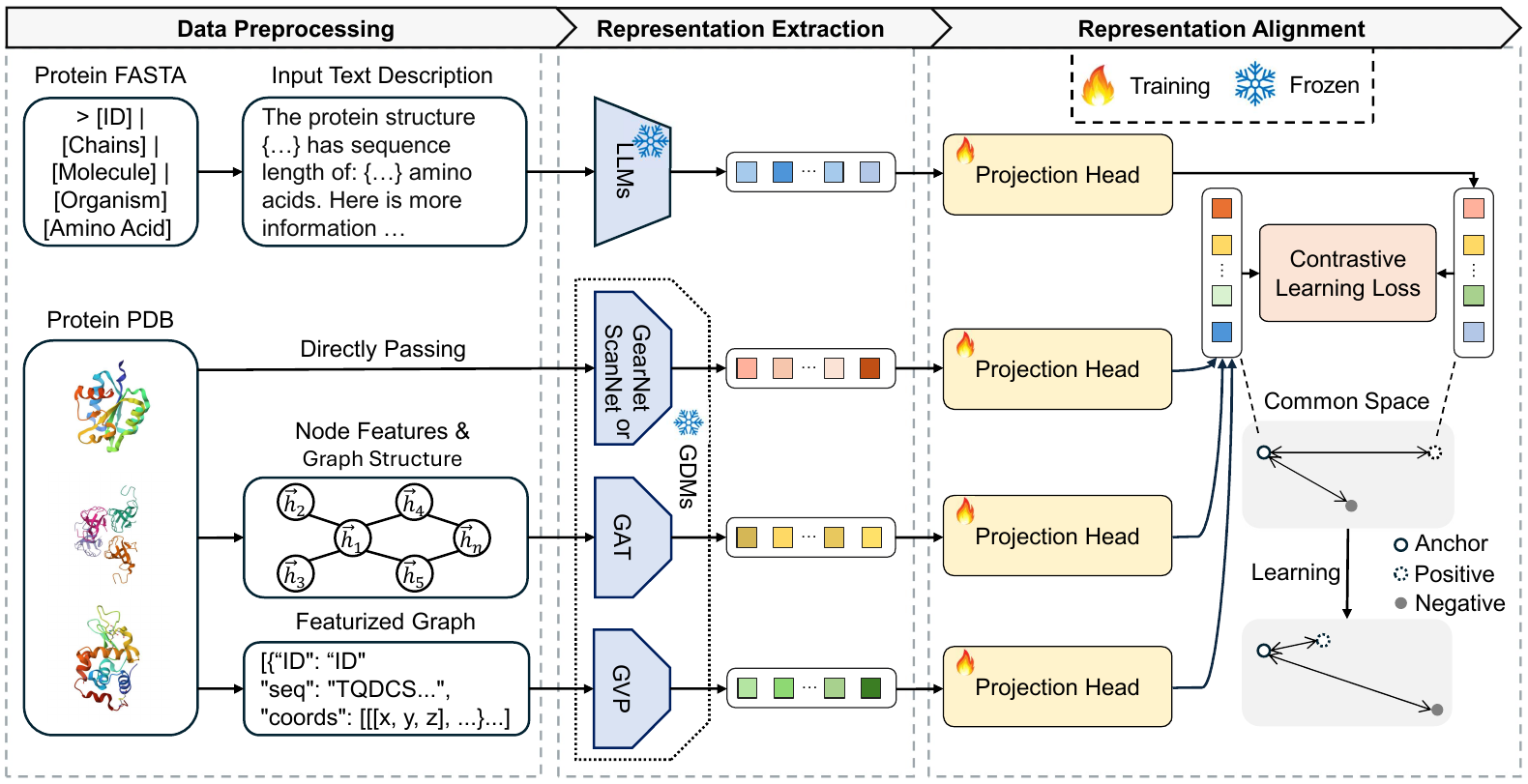}
    \caption{\textbf{Methodology Overview.} The figure is divided into three main parts: Data Preprocessing, Protein Representation Extraction, and Representation Alignment.}
    \label{fig:overview}
\end{figure}

Recently, multimodal large language models (MLLM) such as GPT-4V~\cite{openai2023gpt4v}, Gemini~\cite{team2023gemini}, Llama-3.2~\cite{meta2024llama32}, LLaVA-1.5~\cite{liu2024improved}, etc., have received increasing attention from the community. A MLLM is an LLM-based model with the ability to receive, reason, and output with multimodal information. This extends beyond just text to potentially include various forms of input and output such as image, audio, video, and other sensory data, allowing the model to process and generate various types of information in an integrated manner. In the realm of protein science, specialized MLLMs such as ProtChatGPT~\cite{wang2024protchatgpt}, ProteinChat~\cite{guo2023proteinchat} and ProteinGPT~\cite{xiao2024proteingpt} have emerged as powerful tools for comprehensive protein analysis. These models can process both protein sequences and structures, integrating this information with vast language understanding capabilities. By combining protein sequence and structure encoders with large language models, these systems can perform complex tasks such as property prediction, structure understanding, and even engage in interactive conversations about protein characteristics. This fusion of modalities allows for a more comprehensive and efficient approach to protein analysis, potentially revolutionizing fields such as drug development and biotechnological advancements.

In these protein-focused MLLMs, latent representation alignment serves as a fundamental technique for integrating different types of protein data. A protein can be represented in multiple modalities - as textual descriptions capturing its function and properties, as graph structures representing amino acid relationships, or as 3D coordinate data describing its spatial configuration. While specialized models exist for each of these representations (like LLMs for text and GDMs for structures), the challenge lies in bridging the gap between these different protein representations to enable comprehensive protein analysis. This is where latent representation alignment plays a crucial role, allowing these diverse protein representations to be mapped into a shared space where they can be meaningfully compared and integrated~\cite{wang2024protchatgpt,guo2023proteinchat,xiao2024proteingpt}.

Several techniques are commonly used to build multimodal models, each with distinct methods for integrating information from different modalities. 
In the bio domain, recent models such as Prot2Text \cite{abdine2024prot2text} integrates Graph Neural Networks (GNNs) \cite{wu2020comprehensive} and LLMs using an encoder-decoder architecture to generate detailed free-text descriptions of protein functions. MM-StackEns model \cite{albu2023mm} uses a combination of sequence data processed by a siamese neural network \cite{koch2015siamese} and graph data processed by Graph Attention Networks (GATs) \cite{velickovic2018graph} to perform Protein-Protein Interaction (PPI) prediction \cite{jones1996principles}. ProtST framework \cite{xu2023protst} uses multimodal representation alignment and multimodal mask prediction tasks to align protein sequence representations with biomedical text representations. Among these methods, latent representation alignment remains the most popular due to its simplicity and scalability \cite{duan2022multi}. Protein-focused MLLMs such as ProteinGPT \cite{xiao2024proteingpt} is a multimodal chat system designed for protein analysis, combining protein sequence and structure data with a LLM. This model integrates a protein sequence encoder and structure encoder with a linear projection layer to align diverse embeddings with language model inputs. ProteinChat \cite{guo2023proteinchat} introduces a ChatGPT-like interface for 3D protein structures, emphasizing interactive protein analysis. Its architecture consists of a GVP \cite{jing2020learning} block to encode structural information, a projection layer, and a LLM. The protein embeddings from the GVP are aligned with the LLM’s language representations through the projection layer, allowing detailed question-based responses regarding protein 3D structures. ProtChatGPT \cite{wang2024protchatgpt} introduces a multimodal LLM framework that combines protein sequence and structure data using a specialized Protein-Language Pertaining Transformer (PLP-former) to bridge the gap between protein structures and text data. The PLP-former aligns protein data with textual descriptions, while a projection adapter facilitates multimodal integration with the LLM, allowing for accurate question-based protein analysis. By aligning the latent spaces of different modalities, this technique enables seamless cross-modal integration without requiring complex fusion mechanisms or retraining weights for every feature. 

Despite the current progress in protein-focused MLLMs, latent representation alignment across different protein modalities remains a significant challenge that requires deeper investigation~\cite{xiao2024proteingpt}.  We aim to address these gaps in understanding and provide insights that could lead to more principled and effective approaches in multimodal protein analysis. In the protein domain, LLMs excel at processing and understanding textual descriptions of proteins, including their functions, properties, and relationships documented in scientific literature~\cite{kulmanov2024protein, liu2024plmsearch, hong2024fast, flamholz2024large}. Meanwhile, GDMs are specialized neural architectures designed to capture the complex three-dimensional structural information of proteins, including atomic coordinates, bond angles, and spatial relationships between amino acids~\cite{li2024contextual, mitra2024geometric, he2024highly, zhung20243d}. While both model types offer complementary strengths in protein analysis, effectively aligning their representations remains a significant challenge. Our work explores this alignment challenge through three main areas: I) The model perspective analysis, focusing on alignment performance and correlations between different model pairs. II) The protein perspective analysis, examining which types of proteins align well or poorly across models. III) Strategies for improving alignment, including the impact of GDM representation dimensionality, the effect of projection head complexity, and the benefits of fine-tuning LLMs on protein-specific data. To investigate these areas, we conduct extensive experiments comparing three state-of-the-art LLMs (Gemma2-2B, LLaMa3.1-8B, and LLaMa3.1-70B) with four protein-specialized GDMs (GearNet, GVP, ScanNet, GAT). We analyze the alignment performance of different model pairs, study how protein characteristics affect alignment quality, and evaluate various strategies for enhancing alignment through model architecture modifications and training approaches. An overview of our methodology is provided in Figure \ref{fig:overview}.

Our exploration of the alignment between LLMs and GDMs in the protein domain has revealed several key findings. We found that GDMs integrating both graph and 3D structural information of proteins tend to align better with LLMs. Additionally, larger LLMs with higher embedding dimensions showed improved alignment performance with the same GDM. Our analysis revealed strong correlations between high-alignment model pairs and other high-alignment pairs. Notably, the rarity of a protein significantly affects the model's alignment performance, with rare proteins posing greater challenges. Based on our observations, we highlight the challenges and limitations of current protein datasets for representation alignment, particularly due to unequal levels of study across proteins and the presence of homologous relationships among them. In terms of model architecture, we discovered that retraining GDMs to have higher embedding dimensions enhances alignment with LLMs. The complexity of the projection head also plays a crucial role, as increasing the number of layers improves alignment up to a certain threshold, beyond which benefits diminish. Finally, we found that fine-tuning LLMs with protein-specific data can significantly improve alignment with GDMs. By systematically analyzing how different model architectures, protein characteristics, and alignment strategies affect performance, we can provide concrete guidelines for building protein-focused MLLMs that more accurately capture the relationship between protein structure and function. This improved understanding directly translates to better performance in crucial tasks such as protein function prediction, protein-protein interaction analysis, and drug binding site identification.

\section*{Results}

Our results are organized around six key research questions, which can be organized into three essential perspectives (i.e., Model Perspective, Protein Perspective, and Strategies for Improvement) that comprehensively address the challenges of multimodal alignment in the protein domain. First, the \emph{Model Perspective Analysis} examines which combinations of LLMs and GDMs achieve optimal alignment and how different model pairs correlate, providing fundamental insights into model selection and architecture design. Second, the \emph{Protein Perspective Analysis} investigates the characteristics of proteins that influence alignment quality, crucial for understanding the biological factors that affect model performance and identifying potential limitations in current approaches. Finally, the \emph{Strategies for Improvement} section explores practical methods to enhance alignment performance, including architectural modifications and training techniques, offering concrete solutions to the challenges identified in the previous analyses. Through these three complementary perspectives, we provide a thorough understanding of both the theoretical foundations and practical considerations in protein-focused multimodal alignment.

\subsection*{Evaluation Metric}
In this section, we define key terms for our evaluation. The projected representation refers to the representation after passing through the projection head, while a positive pair represents two modality representations originating from the same protein, and a negative pair represents representations belonging to different proteins. We also define two types of alignment scores: (1) Alignment score between a protein pair, which is the cosine similarity between a projected graph representation and a projected text representation. Ideally, a positive pair should have a higher alignment score, and a negative pair should have a lower alignment score. (2) Alignment score between a model pair, which measures the overall alignment performance.

To calculate the alignment score between a model pair, we measure the alignment of the projected graph representation from a GDM and projected text representation from a LLM. The metric assesses how well the projection heads can map two different modality representations into a common dimensional space. 
Let \( N \) be the number of proteins in the test set, and \( M \) denote the total number of negative pairs. For each protein \( i \), we compute the cosine similarity between the projected graph representation \( g_i \) and the corresponding text representation \( t_i \). The final metric, which quantifies the alignment performance, is defined as the difference between the positive score and the negative score:
\[
\mathcal{F}_{\text{similarity}} = \mathcal{F}_{\text{positive}} - \mathcal{F}_{\text{negative}} = \underbrace{ \frac{1}{N} \sum_{i=1}^{N} \text{sim}(g_i, t_i) }_{{\text{Positive score}}} - \underbrace{ \frac{1}{M} \sum_{i=1}^{N} \sum_{\substack{j=1 \\ j \neq i}}^{N} \left| \text{sim}(g_i, t_j) \right| }_{{\text{Negative score}}}.
\]
Here, the positive score is obtained by averaging these similarities across all \( N \) proteins.
Similarly, for each protein \( i \), we compute the absolute value of the cosine similarity between \( g_i \) and all text representations \( t_j \) where \( j \neq i \), treating these as negative pairs. The negative score is calculated by averaging these values over all negative pairs. Ideally, we expect \( \mathcal{F}_{\text{positive}} \) to be close to 1.0 and \( \mathcal{F}_{\text{negative}} \) to be close to 0.0 for a well-aligned model, aligning with the contrastive loss function designed to constrain cosine similarities within a range of [0, 1]. Details of this contrastive loss function are discussed in the Methods section. A larger \( \mathcal{F}_{\text{similarity}} \) indicates better alignment, where the alignment effectively assigns higher similarities to positive pairs while maintaining low similarities for negative pairs.

\subsection*{Model Perspective Analysis}

\subsubsection*{Model Pairs Alignment (RQ1): Which LLM-GDM model pairs demonstrate the best alignment performance?}

\begin{table}[t]
\footnotesize
\centering
\caption{Model Pairs Alignment Performance (Trained vs Untrained)}
\begin{tabular}{lccc|ccc}
\toprule
\midrule
Models & \multicolumn{3}{c}{Trained} & \multicolumn{3}{c}{Untrained} \\
\cmidrule(lr){2-4} \cmidrule(lr){5-7}
 & Gemma2-2B & LLaMa3.1-8B & LLaMa3.1-70B & Gemma2-2B & LLaMa3.1-8B & LLaMa3.1-70B \\
\midrule
GearNet  & 0.2958 & 0.3211 & 0.3515 & -0.0301 & -0.0345 & -0.0241 \\
GAT      & -0.0533 & -0.0700 & -0.0610 & -0.0001 & -0.0003 & -0.0014 \\
GVP      & 0.0774 & 0.0965 & 0.1408 & -0.0001 & -0.0094 & -0.0252 \\
ScanNet  & 0.3223 & 0.3375 & 0.3856 & -0.0004 & -0.0210 & -0.0094 \\
\midrule
\bottomrule
\label{table:alignment_comparison}
\end{tabular}
\end{table}


We report the results in Table \ref{table:alignment_comparison}. The results indicate that, when no training is applied to the model pairs' projection heads shown on the right of the table, directly mapping the projected graph and text representations results in alignment scores close to zero. This suggests that without any learned mapping between the modalities, there is no meaningful alignment between the models. However, after training the projection heads, shown on the left of the table, the alignment performance improves across almost all model pairs, except for those involving GAT, where the alignment scores remain near zero. Notably, model pairs involving ScanNet or GearNet exhibit significantly higher alignment scores compared to other model pairs. This is because, when processing a protein's 3D structure, both ScanNet and GearNet account not only for the graph structure but also for geometric features, such as the angles between 3D edges, atomic-level features, etc. Another important observation is the positive correlation between the size of the LLM and the alignment performance: as the parameter and embedding dimensionality of the LLMs increase (from Gemma2-2B to LLaMa3.1-70B), the alignment scores also improve. This suggests that larger LLMs, with their higher-capacity embedding spaces, capture richer information about the protein, which in turn enhances alignment performance. Thus, controlling the GDM side of the model pair and increasing the dimension of the LLM leads to better alignment, highlighting the impact of embedding dimensionality on representation quality. 

Recall that the alignment score of a model pair is determined by its positive and negative scores. In Figure \ref{fig:protein_combine_1}A, we track these scores, where the x-axis uses `untrained' to indicate the untrained version of each corresponding GDM paired with Gemma2-2B. As shown, the positive and negative scores for untrained models are close to zero, supporting our earlier observation that there is no meaningful alignment without training. However, after training the projection heads, all model pairs show increases in both positive and negative scores, with most model pairs achieving higher positive scores than negative ones, except for those involving GAT, where the negative scores exceed the positive scores. This indicates that even with trained projection heads, alignment remains ineffective for model pairs with GAT, as both positive and negative scores increase at the similar level. In model pairs with GearNet or ScanNet, we observe that the positive scores continue to rise as the LLM size increases, while the negative scores decrease. This explains why alignment scores improve with larger LLMs. However, this raises a question: `\emph{Why do the negative scores not remain at zero, even though the contrastive loss function is designed to keep them there?}' The answer lies in the way proteins are cataloged and studied. Some proteins have been decomposed and studied multiple times by scientists with different angles, resulting in multiple protein IDs for essentially the same protein. Therefore, even if proteins have different IDs, they may still belong to the same original protein. Additionally, proteins can be homologous, meaning they share structural similarities even if they are technically different proteins. For example, human RNA polymerase II is very similar to yeast RNA polymerase II; despite being distinct proteins with different IDs, they exhibit similarities.

These findings reveal significant challenges when aligning multimodal models in the protein domain. First, purely aligning proteins based on matching IDs or names may not be sufficient due to the nuanced relationships between proteins. Second, when the contrastive loss function forces the similarity of negative homologous protein pairs toward zero, it may also inadvertently reduce the similarity of positive pairs, offsetting the decrease intended for negative pairs. This suggests that both positive and negative pairs' similarities may decrease, complicating effective alignment. Addressing these challenges requires a more sophisticated approach that accounts for homologous and structurally related proteins.

\subsubsection*{Pearson Correlation (RQ2): Is there a correlation between different model pairs?} 
 
\begin{figure}[t]
    \centering
\includegraphics[width=0.9\linewidth]{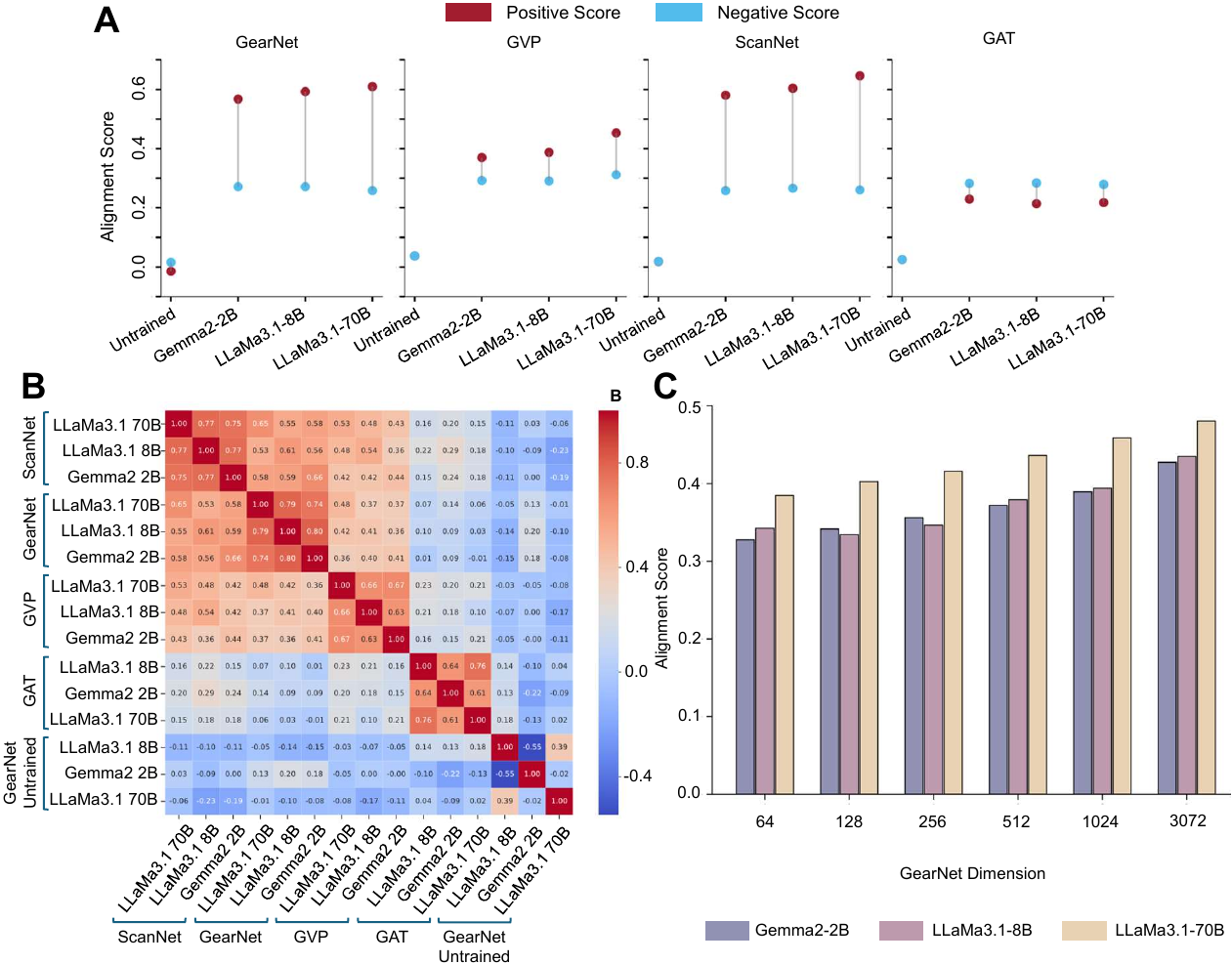}
    \caption{\textbf{Model Perspective Analysis.} (A) Alignment Score Details. The figure displays the positive (red dot) and negative (blue dot) alignment scores for each model pair. The y-axis represents the alignment score, while the x-axis indicates the different LLMs used in the model pairs. The difference between the positive and negative score is the model pair alignment performance. (B) Pearson Correlation Coefficient. The heatmap shows the Pearson correlation coefficients between different model pairs, with model pairs on the x-axis compared against those on the y-axis. (C) GDM Dimension Analysis. It shows the effect of increasing the GearNet embedding dimension (x-axis) on the alignment score (y-axis) for different model pairs.}
    \label{fig:protein_combine_1}
\end{figure}

We computed the Pearson Correlation Coefficient for all trained model pairs, plus the untrained model pairs that have GearNet on the GDM side.
For each protein in the testing set, we first obtain its corresponding GDM and LLM representations. We then calculate the alignment score between these two projected representations for each model pair. This process is repeated across all model pairs, resulting in a list of proteins where each protein is associated with alignment scores from different model pairs. Finally, we analyze these alignment scores to compute the correlation between different model pairs.
The results, presented in Figure \ref{fig:protein_combine_1}B, reveal several interesting patterns.

First, the untrained version of the model pairs shows almost no correlation with any of the trained model pairs, as indicated by the near-zero correlation values. This suggests that without training the projection heads, the latent representations from different modalities do not align, which also cause no alignments between model pairs. This finding is consistent with the findings from the Research Question 1.
Another interesting observation is that the untrained model pair of GearNet with LLaMa3.1-8B shows relatively high correlation score of 0.39 with the untrained model pair of GearNet with LLaMa3.1-70B. This suggests that, as both LLMs are part of the LLaMa3.1 family, their latent representations are closely related, allowing them to represent information in a similar manner. In contrast, the untrained model pair of GearNet with LLaMa3.1-8B exhibits a strong negative correlation with the untrained model pair of GearNet with Gemma2-2B. This indicates that while LLaMa3.1-8B and LLaMa3.1-70B may encode information similarly, LLaMa3.1-8B and Gemma2-2B may represent information in fundamentally different, or even opposing ways.
Secondly, among the trained model pairs, model pairs involving ScanNet or GearNet exhibit the highest correlation, suggesting a strong similarity of these two model pairs when aligning different modalities representations of a protein. In contrast, the GAT-based model pair consistently show low correlation with all other pairs, indicating that the GAT-based model pair aligns differently and less effectively compared to other model pairs.
Moreover, when examining the correlation between two trained model pairs that share the same GDMs, we observe consistently high correlation across all model pairs, regardless of the LLM used. This finding contrasts with the untrained model pairs, where different LLM families might represent information differently. After training, however, the representations of all LLMs become more aligned, indicating that the training process effectively harmonizes the embeddings across different LLMs, allowing them to represent information in a similar manner.

These findings align with the results from the first research question, where model pairs involving ScanNet or GearNet achieved the highest alignment scores. The strong correlation between high-alignment model pairs suggests that if a protein’s modalities can be highly aligned in one model pair, they are likely to align similarly in other high-performance model pairs as well. This indicates a shared underlying structure in the latent spaces of GearNet and ScanNet, which makes them more suitable for multi-modal alignment with LLMs.

\begin{figure}[ht]
    \centering
    \includegraphics[width=0.97\linewidth]{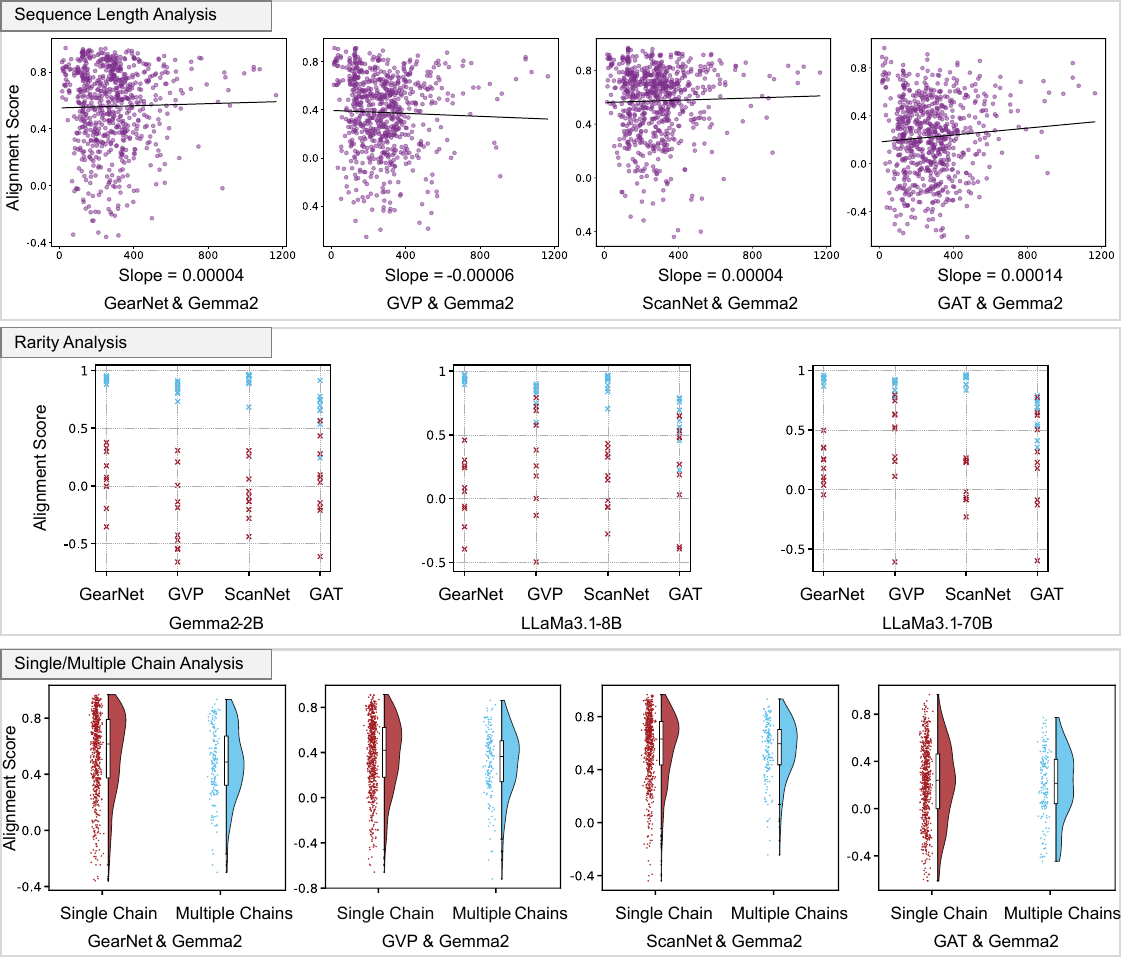}
    \caption{\textbf{Protein Perspective Analysis.} At the top, the figure illustrates whether a protein's amino acid sequence length affects its alignment score across different model pairs. In the middle, the figure illustrates whether rarity of a protein affects its alignment score across different model pairs. At the bottom, the figure illustrates whether the number of chains (single or multiple) affects the alignment score across different model pairs, as indicated below each graph. }
    \label{fig:protein_analysis}
\end{figure}

\subsection*{Protein Perspective Analysis}

\subsubsection*{RQ3: What types of proteins align well across all model pairs, and which do not?}

To analyze the alignment of different proteins, we define three key properties of proteins: amino acid sequence length, rarity, and single/multiple chains. Within each model pair, we calculate the alignment score between the projected graph and text representations for each protein.

\vspace{5pt}
\noindent
\emph{Factor 1: Amino Acid Sequence Length:}
When analyzing the effect of amino acid sequence length, we conducted experiments on model pairs with Gemma2-2B as LLM side. We categorize the proteins based on their amino acid sequence length. As shown in Figure \ref{fig:protein_analysis}a, the x-axis represents the sequence length, and the y-axis represents the alignment score. Each dot corresponds to a protein with a particular sequence length and its corresponding alignment score from the model pair. We applied simple linear regression to measure the correlation between sequence length and alignment score, which is represented by the trend line. The results indicate that sequence length has no significant impact on alignment, as the slope of the regression line is near zero.

\vspace{5pt}
\noindent
\emph{Factor 2: Rarity:}
Determining rarity of a protein is more complex. We first extract the protein's molecular name and organism, then search for the frequency of similar proteins belonging to the same molecule and organism. Proteins with numerous occurrences are classified as `popular,' while those with few or no similar occurrences are classified as `rare.' For instance, a rare protein such as `3I1A' belongs to `Spectinomycin phosphotransferase, Legionella pneumophila,' a combination that is hard to find in other proteins. In contrast, a popular protein like `3PYK' belongs to `Carbonic anhydrase, Homo sapiens.'

We randomly selected 10 popular and 10 rare proteins for analysis and found that rarity significantly affects alignment performance. The details of the random selection and these 20 proteins are provided in Supplemental Note \ref{appendix:protein}. As shown in Figure \ref{fig:protein_analysis}b, the x-axis represents different model pairs, while the y-axis represents alignment scores. Each dot corresponds to a protein’s alignment score for a specific model pair. Blue dots represent popular proteins, and red dots represent rare proteins. The results clearly show that popular proteins tend to align better, while rare proteins generally show poorer alignment. This observation further supports our proposed challenge in Research Question 1: current protein datasets are often unbalanced, with certain proteins extensively studied by multiple researchers and assigned different IDs despite being the same protein, while others are studied only once. This imbalance explains why popular proteins align better than rare ones, as popular proteins are encountered more frequently during training. Even though most model pairs struggle to align rare proteins, ScanNet- and GearNet-based pairs manage to maintain alignment scores closer to zero, adhering to the behavior expected from the contrastive loss function. In contrast, GVP- and GAT-based pairs exhibit a more scattered distribution of alignment scores, with GAT in particular showing poor alignment for both popular and rare proteins. 
Another noteworthy finding is that as the size of the LLMs increases, the distribution of proteins' alignment scores becomes more compact. This is particularly evident in model pairs such as GearNet with Gemma2-2B compared to GearNet with LLaMa3.1-70B, and similarly for ScanNet paired with Gemma2-2B versus ScanNet with LLaMa3.1-70B. In both cases, the alignment scores for popular proteins show a tighter distribution, and this trend also holds for rare proteins. These results are consistent with our previous findings, further emphasizing that model pairs involving ScanNet or GearNet demonstrate better overall alignment and share stronger correlations. Additionally, larger LLMs contribute to better alignment performance, highlighting the importance of embedding size and model capacity in improving multimodal alignment.

\vspace{5pt}
\noindent
\emph{Factor 3: Single/Multiple Chains:}
We also investigated whether the number of chains in a protein, as derived from the FASTA file, correlates with alignment performance. As shown in Figure \ref{fig:protein_analysis}c, 
the x-axis represents single or multiple chains, while the y-axis represents the alignment score. The violin plots represent the distribution of alignment scores for single and multiple chain proteins, with the width of the violin indicating the density of data points at different alignment score levels. The scatter points (red for single-chain and blue for multiple-chain proteins) show individual protein alignment scores, while the boxplots within the violins depict the median, interquartile range, and overall spread of the alignment scores.
The results show no clear relationship between the number of chains and alignment performance.

\vspace{5pt}
\noindent
In summary, our analysis shows that of the three protein properties examined, rarity has the most significant impact on alignment performance. Specifically, model pairs tend to achieve higher alignment scores with popular proteins and lower scores with rare proteins. This insight suggests that when developing multi-modal models, special attention should be paid to rare proteins, as they are more challenging to align effectively.

\subsection*{Strategies for improvement}

\subsubsection*{GDM Dimension Analysis (RQ4): Does increasing GDM dimension improve alignment?}

From Research Question 1, we reveal that if we control the GDM side of the model pair, and increase the parameter and dimension of the LLM, the model pair will show better alignment. In this section, we control the GDM side to GearNet model. Then, we retrained the GearNet with different size of hidden layer dimension. We aim to investigate how the dimensional size of GearNet influences the alignment between the model pairs. In our paper, we retrain the GearNet so that its output dimension will be \{64, 128, 256, 512, 1024, 3072\}. Different from the previous experiment that directly use the pre-trained GearNet, our retrained task for the GearNet is Multiple Binary Classification.

Multiple Binary Classification is a task in which several binary classification problems are solved simultaneously. A protein may possess several distinct functional properties, each of which can be treated as a separate binary classification problem. Formally, given a dataset of \( n \) proteins \( \{ (x_i, y_i) \}_{i=1}^n \), where \( x_i \in \mathcal{X} \) represents the input features (e.g., protein graph representations) and \( y_i \in \{0, 1\}^k \) denotes the binary labels for \( k \) different tasks (functions), the objective is to learn a function \( f : \mathcal{X} \to \{0, 1\}^k \) that predicts a binary label for each task. Each task \( j \in \{1, 2, \dots, k\} \) is a binary classification problem, where the predicted probability \( \hat{y}_{i,j} \in [0, 1] \) represents the likelihood that the \( i \)-th protein possesses the \( j \)-th function.
The learning objective is defined by minimizing a binary cross-entropy loss across all tasks:
\[
L = - \frac{1}{n} \sum_{i=1}^n \sum_{j=1}^k \left[ y_{i,j} \log(\hat{y}_{i,j}) + (1 - y_{i,j}) \log(1 - \hat{y}_{i,j}) \right],
\]
where \( y_{i,j} \) is the ground truth label and \( \hat{y}_{i,j} \) is the predicted probability for task \( j \) and protein \( i \). In this way, the model is trained to simultaneously predict multiple binary outcomes, which can capture the presence or absence of different properties or functions in a single sample, such as the functions associated with a particular protein. The dataset used for training is officially provided by TorchDrug \cite{zhu2022torchdrug}, a comprehensive machine learning library designed for drug discovery and protein analysis. The dataset consists of three subsets: the training set contains 15,170 samples, the validation set comprises 1,686 samples, and the test set includes 1,860 samples. Each sample in the dataset is annotated with multiple binary labels corresponding to different protein functions. The implementation detail is shown in Supplemental Note \ref{appendix:gearnet_train}.

After retraining the GearNet model, we conducted the experiment outlined in Research Question 1. This involved training the projection head for each side of the model pair and evaluating the performance using our defined metric.
As illustrated in Figure \ref{fig:protein_combine_1}C, there is a clear trend indicating that as the dimensionality of the GearNet’s embeddings increases, the alignment score of the model pairs improves. This pattern is consistent regardless of which LLM is used. Additionally, we observe that model pairs with larger LLMs, such as LLaMa3.1-70B, tend to achieve higher alignment scores. An interesting observation is that when the GearNet embedding dimension is 256 or smaller, model pairs with Gemma2-2B and LLaMa3.1-8B perform comparably, with each occasionally outperforming the other. However, once the GearNet dimension exceeds 512, model pairs with LLaMa3.1-8B consistently outperform those with Gemma2-2B.
This result is consistent with our findings in Research Question 1, where we observed that as the LLM's parameters increased, its representation captured richer information, leading to better alignment. Similarly, in this experiment, the larger GearNet dimensions introduce richer, more detailed representations of the protein graph, resulting in enhanced alignment scores. This suggests that increasing the complexity of the GDM, like increasing the parameters in the LLM, helps the model capture more intricate biological information.
\begin{figure}[t]
    \centering
    \includegraphics[width=0.8\linewidth]{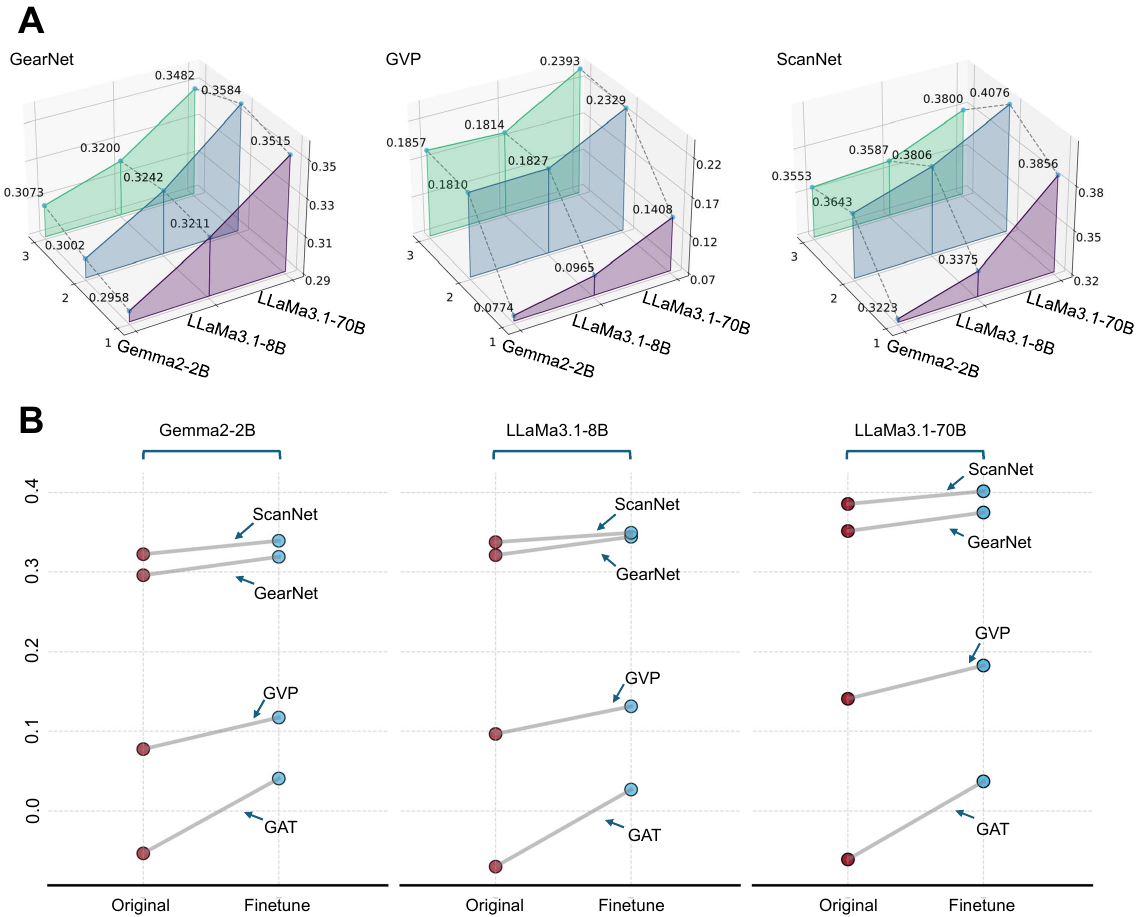}
    \caption{\textbf{Strategies for Improvement.}
    (A) The figure consists of three 3D graphs: model pairs with GearNet are shown on the left, model pairs with GVP in the middle, and model pairs with ScanNet on the right. Each graph has three axes: the x-axis (horizontal axis in the foreground) represents the LLM used in the model pair, the y-axis (horizontal axis on the left) represents the number of layers in the projection head, and the z-axis (vertical axis) represents the alignment score. (B) The figure is divided into three sections: Gemma2-2B on the left, LLaMa3.1-8B in the middle, and LLaMa3.1-70B on the right. The y-axis represents the alignment score, and the x-axis indicates whether the LLM is fine-tuned or not. Each line represents the trend in alignment scores for a specific model pair, from the original (pre-fine-tuning) to post-fine-tuning.}
    \label{fig:protein_combine_2}
\end{figure}

\subsubsection*{Projection Head Analysis (RQ5): Does adding layers to the GDM's projection head enhance alignment performance?}

In previous experiments, all projection heads consisted of a single linear layer. In this section, we explore whether increasing the complexity of the GDM's projection head by adding more layers improve alignment performance.
To investigate this, we added one and two additional linear layers to the GDM's projection head while keeping the LLM's projection head structure unchanged. This results in GDM projection heads with 2 or 3 linear layers, with ReLU activation functions between layers. We then re-trained the projection heads and evaluated alignment performance as in previous experiments. Further details of the implementation are provided in Supplemental Note \ref{appendix:proj_head_layer}.

As shown in Figure \ref{fig:protein_combine_2}A, the x-axis represents different model pairs, the y-axis represents the number of layers, and the z-axis represents the alignment score. The results indicate that increasing the number of layers in the projection head consistently improves alignment scores for the majority of model pairs. Notably, projection heads with 2 layers significantly outperform the 1-layer architecture across all model pairs. However, adding a third layer provides only marginal improvements and, in most cases, actually decreases the alignment score, especially for model pairs involving ScanNet.
The improvement from 1 to 2 layers is substantial, suggesting that a 2-layer architecture provides a significant boost in alignment capability. However, the minimal gains observed when moving from 2 to 3 layers indicate diminishing returns with increased complexity. Thus, the 2-layer architecture appears to strike the best balance between model complexity and performance improvement. This suggests that the number of projection head layers should be carefully determined based on empirical validation for each specific model combination, rather than assuming a one-size-fits-all approach. Figure \ref{fig:proj_head_detail}B shows the continuous graph analysis of model pairs with GAT due to the page limit.

\subsubsection*{LLM Fine-Tuning Analysis (RQ6): Does fine-tuning an LLM on protein data enhance alignment performance?}

In previous experiments, all model pairs used pre-trained LLMs directly downloaded from Huggingface. In this section, we fine-tuned all LLMs on the dataset constructed during the FASTA preprocessing stage, where we generated a detailed text description containing all relevant information for each protein. The fine-tuning process involved framing the task as a question-answering problem, where the input was a prompt such as, “What is the protein: XXX?” and the expected output was the corresponding text description. Further implementation details are provided in Supplemental Note \ref{appendix:llm_finetune}.
After fine-tuning the LLMs, we trained the 1-layer projection heads for the each model pairs and evaluated the alignment performance as before. As shown in Figure \ref{fig:protein_combine_2}B, the results clearly show that fine-tuning improves alignment scores across all model pairs. This indicates that providing the LLM with domain-specific knowledge enhances its ability to align with GDM representations.
These findings suggest that LLMs fine-tuned on domain-specific data are more capable of generating representations that align effectively with other modalities. This underscores the importance of domain-specific LLMs when developing multimodal models, as they facilitate better alignment with non-text modalities such as graphs.

\section*{Discussion}

In this paper, we investigated the alignment between LLMs and GDMs in the protein domain through six research questions, focusing on mapping their respective representations into a common space defined by LLM embedding dimensions. Our comprehensive analysis revealed key insights into effective alignment strategies and produced concrete recommendations for designing future protein-based multimodal LLMs.

\subsection*{Alignment Observations and Takeaways}

Our research questions can be broadly divided into three categories. Research Questions 1-2 focused on model perspective analysis. We explored which pairs of LLMs and GDMs achieve better alignment. We found that GDMs capable of learning not only from the graph structure of proteins but also from their 3D geometric information tend to align more effectively with LLMs. Additionally, LLMs with higher embedding dimensions consistently demonstrated better alignment with GDMs. We also observed that model pairs with strong alignment tend to have higher correlation with other high-performing model pairs. Our analysis also uncovered significant challenges when aligning multimodal models in the protein domain. One major issue is that different protein IDs may represent different perspectives or angles of the same underlying protein, complicating alignment efforts. Additionally, proteins are often homologous, meaning that even distinct proteins can share structural and functional similarities. On a more granular level, Research Question 3 focused on protein perspective analysis, we discovered that popular proteins are generally easier to align across model pairs, whereas rare proteins present greater challenges in achieving alignment.
Research Questions 4-6 examined methods for improving alignment between model pairs. Our findings indicate that increasing the embedding dimension size of a GDM enhances its alignment with LLMs. Moreover, adding an extra linear layer to the GDM’s projection head can significantly improve alignment performance, though this improvement stopped after two layers, suggesting diminishing returns with further complexity. Another key insight is that fine-tuning LLMs with protein-specific knowledge can lead to better alignment with GDMs, highlighting the value of domain-specific adaptation.

Our analysis also revealed several fundamental limitations in current protein databases that pose significant challenges for multimodal alignment. First, the imbalance in protein documentation presents a major obstacle. Protein databases often exhibit biases, with certain proteins being studied and documented multiple times under different IDs, while others are rarely studied or only appear once. This over-representation of well-studied proteins leads to models that align more effectively with these familiar proteins but struggle with rare or underrepresented ones. Therefore, biologists should also focus more on studying rare proteins to ensure that all proteins have a similar level of documentation and representation quality. Second, the biological reality of protein homology creates inherent complexity, where distinct proteins share structural and functional similarities due to common evolutionary origins. This makes it difficult to determine appropriate similarity thresholds, as even different proteins may share similar features in sequence and structure. Third, we discovered a notable dataset imbalance where popular, well-studied proteins generally achieve better alignment scores across model pairs compared to rare proteins. This points to a fundamental sampling bias in protein databases, where frequently studied proteins are overrepresented while rare proteins lack sufficient documentation for effective alignment. These database limitations highlight the need for more sophisticated alignment approaches that can account for protein relationships beyond simple ID matching, as well as methods to address the inherent imbalances in protein documentation and representation.

\subsection*{Protein-focused MLLMs Design Recommendations}
Our findings from Research Questions 4-6 provide important insights for the design of future multimodal LLMs in the protein domain. We discovered that GDMs which incorporate both graph and 3D structural information of proteins demonstrate superior alignment with LLMs, suggesting future designs should prioritize models capable of capturing these multidimensional protein representations. The complex relationships between different proteins must be carefully considered during the alignment process to avoid oversimplified mappings that fail to capture subtle protein interactions. Our research also revealed that higher-dimensional embeddings in LLMs contribute to better alignment by capturing richer semantic information, indicating that increasing the capacity of the LLM's embedding space can enhance overall performance. When designing projection heads, we found that a two-layer architecture offers the optimal balance between simplicity and performance, as additional layers provide diminishing returns. Furthermore, fine-tuning LLMs with domain-specific data, such as protein descriptions, significantly improves alignment with GDMs, highlighting the importance of customizing LLMs to better understand the intricacies of the protein domain for more effective cross-modal integration.

\begin{table}[t]
    \centering
    \caption{Protein-focused MLLMs Performance on Protein Description Task}
    \begin{tabular}{lcc|cc|cc}
        \toprule
        \midrule
        Models & \multicolumn{2}{c}{Gemma2-2B} & \multicolumn{2}{c}{LLaMa3.1-8B} & \multicolumn{2}{c}{LLaMa3.1-70B} \\ \cmidrule(lr){2-3} \cmidrule(lr){4-5} \cmidrule(lr){6-7}
        & ROUGE & BLEU & ROUGE & BLEU & ROUGE & BLEU \\
        \midrule
        GearNet     & 0.3761 & 0.1914 & 0.4164 & 0.2125 & 0.4849 & 0.2574 \\
        GAT         & 0.1256 & 0.0903 & 0.1278 & 0.0916 & 0.1315 & 0.1009 \\ 
        GVP         & 0.1364 & 0.0952 & 0.1420 & 0.0978 & 0.1536 & 0.1058 \\ 
        ScanNet     & 0.4203 & 0.2057 & 0.4614 & 0.2392 & 0.5127 & 0.2620 \\
        \midrule
        \bottomrule
    \end{tabular}
    \label{tab:model_performance}
\end{table}

\subsection*{Protein-focused MLLMs Performance and Applications}

We have explored the alignment landscape between LLMs and GDMs, proposing several strategies to enhance their alignment. However, a critical question arises: `How does this alignment contribute to protein modeling or real-world applications?' To address this, we developed protein-focused MLLMs by aligning each LLM-GDM model pair and further fine-tuning the projection head within the protein-focused MLLM. Existing research has demonstrated that the degree of alignment in multimodal models is strongly correlated with their performance on specific downstream tasks. Additionally, alignment levels play a significant role in mitigating hallucinations, where models generate plausible but factually incorrect information \cite{sahoo2024comprehensive, 10.1145/3703155}. To evaluate the impact of alignment on both performance and hallucination, we selected the protein description task and employed ROUGE \cite{lin-2004-rouge} and BLEU \cite{papineni2002bleu} metrics. Detailed methodology for building protein-focused MLLMs and example outputs are provided in Supplemental Note \ref{appendix:protein_llm}.

As shown in Table \ref{tab:model_performance}, protein-focused MLLMs utilizing GAT and GVP as their GDMs exhibit poor performance on the protein description task. In contrast, protein-focused MLLMs that employ GearNet and ScanNet as their GDMs achieve superior results. Additionally, we observe a significant improvement in performance as the underlying LLM becomes more powerful (e.g., from Gemma2-2B to LLaMa3.1-70B). This finding aligns with our results from Research Question 1. Furthermore, ROUGE scores are generally higher than BLEU scores. This discrepancy arises because BLEU emphasizes precision—measuring how much of the generated output matches the ground truth—whereas ROUGE focuses on recall, which captures how much of the ground truth is reflected in the generated output. Upon examining the outputs generated by protein-focused MLLMs (see Figure \ref{fig:protein-LLM}b), we discovered that models often generate additional protein-related knowledge not included in our ground truth. These extra details, while potentially accurate, are not reflected in the ground truth, resulting in lower BLEU scores compared to ROUGE. This highlights the need for a more comprehensive protein description benchmark that encompasses a broader and more diverse range of protein knowledge than ours, paving the way for more robust evaluations in future research.

To further illustrate the relationship between alignment levels and hallucination, we present examples of protein-focused MLLMs generated descriptions in Figure \ref{fig:protein-LLM}b for a popular protein ID 4NWH. When the model's alignment is low, most of the generated content consists of hallucinations. This occurs because the model fails to align the protein representation with the actual protein, instead aligning it with unrelated concepts. For example, models using GAT and GVP as GDMs generate hallucinated and irrelevant content, such as `The protein is a protein...' or `It is a large molecule made up of long chains of amino acids.' Although these statements are technically correct, they lack meaningful insights and are as unhelpful as hallucinated content. In contrast, as alignment improves, the level of hallucination decreases significantly. This improvement is evident when comparing models that employ GearNet or ScanNet as GDMs and LLaMa3.1-70B as the LLM to those with lower alignment levels. These findings validate our experimental results and corroborate existing research, demonstrating a strong correlation between higher alignment, better downstream task performance, and reduced hallucination.

\begin{table}[t]
\centering
\caption{Model Pairs Alignment Performance (Reweight)}
\begin{tabular}{lccc}
\toprule
\midrule
Models & \multicolumn{3}{c}{Trained}  \\
\cmidrule(lr){2-4}  
 & Gemma2-2B & LLaMa3.1-8B & LLaMa3.1-70B   \\
\midrule
GearNet  & 0.2958\textit{+0.0260} & 0.3211\textit{+0.0425} & 0.3515\textit{+0.0491}     \\
GAT      & -0.0533\textit{+0.0048} & -0.0700\textit{+0.0282} & -0.0610\textit{+0.0164} \\
GVP      & 0.0774\textit{+0.0192} & 0.0965\textit{+0.0051} & 0.1408\textit{+0.0120} \\
ScanNet  & 0.3223\textit{+0.0367} & 0.3375\textit{+0.0478} & 0.3856\textit{+0.0496} \\
\midrule
\bottomrule
\label{table:alignment_reweight}
\end{tabular}
\end{table}

\subsection*{Rare Protein Alignment Strategies}

In this section, we address the limitations and challenges posed by current protein datasets, particularly the imbalance in the study of popular versus rare proteins. To tackle this rare protein challenge, we propose solutions from two perspectives: improving model-pair alignment and enhancing the downstream task performance of protein-focused MLLMs.

To improve model-pair alignment, we employ a reweighting technique during the training of the projection head. This technique prioritizes rare proteins by assigning higher penalties to errors associated with them. Detailed information on this method can be found in Supplemental Note \ref{appendix:reweighting}. Beyond alignment, we introduce a retrieval-augmented approach to enhance protein-focused MLLMs' performance on the protein description task. Using the reweighted projection head, we follow the methodology described in Supplemental Note \ref{appendix:protein_llm} to construct the protein-focused MLLM. During input stage, our method retrieves the top $k$ most similar proteins to the input protein and augments the original input with this retrieved information. Further details on the retrieval method are provided in Supplemental Note \ref{appendix:retrieval}.

Table \ref{table:alignment_reweight} presents the alignment scores of model pairs after applying the reweighting technique. Original scores from Table \ref{table:alignment_comparison} are shown in black, while improvements are in \textit{italic}. These improvements demonstrate the effectiveness of reweighting in addressing the imbalance between rare and popular proteins. While our classification of rare and popular proteins was not completely, our results suggest that solving the rare protein problem is a promising direction for improving model-pair alignment. Future research should focus on creating a more comprehensive dataset that explicitly distinguishes between rare and popular proteins, enabling further advancements in alignment strategies.

Table \ref{tab:performance_before_after} compares the downstream performance of protein-focused MLLMs before and after applying the retrieval method. Both the reweighting technique alone and the combination of reweighting with retrieval show practical improvements. However, the gains from reweighting alone are limited, as the technique only targets a subset of rare proteins during training. Consequently, a significant portion of rare proteins is penalized similarly to popular proteins. This finding underscores the need for a more detailed classification of rare and popular proteins to optimize reweighting strategies. In contrast, the retrieval method demonstrates significant performance gains for protein-focused MLLMs, particularly with $k=3$ and $k=5$. Figure \ref{fig:reweight_retrieval} illustrates example outputs for rare protein 3I1A before and after applying retrieval with $k=3$. While reweighting improves overall model-pair alignment, it does not fully mitigate hallucinations for rare protein inputs. However, after applying the retrieval method, hallucination rates are substantially reduced, even when compared to the outputs for popular proteins shown in Figure \ref{fig:protein-LLM}b. Moreover, the format and structure of the generated descriptions align more closely with our ground truths, explaining the observed improvements in ROUGE and BLEU scores. These results highlight the importance of a comprehensive ground truth for protein descriptions, as model performance, retrieval augmentation, and evaluation metrics all heavily depend on its quality.

On the other hand, increasing $k$ to 10 results in decreased model performance. We attribute this behavior to the retrieval algorithm retrieving less similar proteins as $k$ increases. These less similar proteins introduce noise or conflicting information, which negatively impacts model performance.



\begin{table}[t]
    \centering
    \scriptsize
    \caption{Protein-focused MLLMs Performance Before and After Retrieval}
    \begin{tabular}{lccc|cc|cc|cc}
        \toprule
        \midrule
        & \multicolumn{1}{c}{} & \multicolumn{2}{c}{Gemma2-2B} & \multicolumn{2}{c}{LLaMa3.1-8B} & \multicolumn{2}{c}{LLaMa3.1-70B} \\ \cmidrule(lr){3-4} \cmidrule(lr){5-6} \cmidrule(lr){7-8}
        & model & ROUGE & BLEU & ROUGE & BLEU & ROUGE & BLEU \\
        \midrule
        \multirow{4}{*}{$reweight$} 
        & GearNet     & 0.3761\textit{+0.0108} & 0.1914\textit{+0.0049} & 0.4164\textit{+0.0141} & 0.2125\textit{+0.0087} & 0.4849\textit{+0.0255} & 0.2574\textit{+0.0074} \\
        & GAT         & 0.1256\textit{+0.0006} & 0.0903\textit{+0.0002} & 0.1278\textit{+0.0014} & 0.0916\textit{+0.0001} & 0.1315\textit{+0.0012} & 0.1009\textit{+0.0008} \\ 
        & GVP         & 0.1364\textit{+0.0079} & 0.0952\textit{+0.0014} & 0.1420\textit{+0.0031} & 0.0978\textit{+0.0005} & 0.1536\textit{+0.0046} & 0.1058\textit{+0.0010} \\ 
        & ScanNet     & 0.4203\textit{+0.0133} & 0.2057\textit{+0.0053} & 0.4614\textit{+0.0174} & 0.2392\textit{+0.0067} & 0.5127\textit{+0.0247} & 0.2620\textit{+0.0081} \\
        \midrule
        \multirow{4}{*}{$k=3$} 
        & GearNet     & 0.3761\textit{+0.1439} & 0.1914\textit{+0.3167} & 0.4164\textit{+0.1343} & 0.2125\textit{+0.3128} & 0.4849\textit{+0.0985} & 0.2574\textit{+0.3207} \\
        & GAT         & 0.1256\textit{+0.0044} & 0.0903\textit{+0.0131} & 0.1278\textit{+0.0061} & 0.0916\textit{+0.0227} & 0.1315\textit{+0.0047} & 0.1009\textit{+0.0141} \\ 
        & GVP         & 0.1364\textit{+0.0408} & 0.0952\textit{+0.0819} & 0.1420\textit{+0.0381} & 0.0978\textit{+0.0742} & 0.1536\textit{+0.0360} & 0.1058\textit{+0.0654} \\ 
        & ScanNet     & 0.4203\textit{+0.1231} & 0.2057\textit{+0.3006} & 0.4614\textit{+0.1138} & 0.2392\textit{+0.3341} & 0.5127\textit{+0.1280} & 0.2620\textit{+0.3565} \\
        \midrule
        \multirow{4}{*}{$k=5$} 
        & GearNet     & 0.3761\textit{+0.1697} & 0.1914\textit{+0.3418} & 0.4164\textit{+0.1557} & 0.2125\textit{+0.3374} & 0.4849\textit{+0.1183} & 0.2574\textit{+0.3443} \\
        & GAT         & 0.1256\textit{+0.0071} & 0.0903\textit{+0.0147} & 0.1278\textit{+0.0068} & 0.0916\textit{+0.0253} & 0.1315\textit{+0.0052} & 0.1009\textit{+0.0169} \\ 
        & GVP         & 0.1364\textit{+0.0536} & 0.0952\textit{+0.0861} & 0.1420\textit{+0.0579} & 0.0978\textit{+0.0829} & 0.1536\textit{+0.0415} & 0.1058\textit{+0.0834} \\ 
        & ScanNet     & 0.4203\textit{+0.1678} & 0.2057\textit{+0.3519} & 0.4614\textit{+0.1492} & 0.2392\textit{+0.3506} & 0.5127\textit{+0.1497} & 0.2620\textit{+0.3904} \\
        \midrule
        \multirow{4}{*}{$k=10$} 
        & GearNet     & 0.3761\textit{+0.1127} & 0.1914\textit{+0.2491} & 0.4164\textit{+0.1136} & 0.2125\textit{+0.2382} & 0.4849\textit{+0.0902} & 0.2574\textit{+0.2965} \\
        & GAT         & 0.1256\textit{+0.0036} & 0.0903\textit{+0.0124} & 0.1278\textit{+0.0054} & 0.0916\textit{+0.0188} & 0.1315\textit{+0.0031} & 0.1009\textit{+0.0136} \\ 
        & GVP         & 0.1364\textit{+0.0333} & 0.0952\textit{+0.0616} & 0.1420\textit{+0.0483} & 0.0978\textit{+0.0709} & 0.1536\textit{+0.0375} & 0.1058\textit{+0.0713} \\ 
        & ScanNet     & 0.4203\textit{+0.0910} & 0.2057\textit{+0.2743} & 0.4614\textit{+0.1028} & 0.2392\textit{+0.3180} & 0.5127\textit{+0.0844} & 0.2620\textit{+0.3521} \\
        \midrule
        \bottomrule
    \end{tabular}
    \label{tab:performance_before_after}
\end{table}


\section*{Conclusions}

In this paper, we conducted a comprehensive analysis of multimodal alignment between LLMs and GDMs in the protein domain, addressing six key research questions. Our study analyzed the models alignment and proteins alignment, such as the types of models that align well together, and specific factors that influence alignment performance, such as protein rarity and model complexity. Through these experiments, we uncovered significant challenges in multimodal alignment and provided valuable insights into its underlying mechanics. Our findings offer practical guidelines for designing more effective multimodal models. We highlighted the importance of accounting for nuanced and homologous relationships between proteins, emphasizing that simple mappings may overlook critical similarities. We demonstrated the importance of leveraging 3D geometric information in GDMs, maximizing embedding dimensions in LLMs, and fine-tuning LLMs with domain-specific knowledge. Additionally, we identified optimal layers for projection heads to enhance alignment without unnecessary complexity. Furthermore, we translated the alignment between LLMs and GDMs into practical applications by developing protein-focused MLLMs and evaluating them on the protein description task. The downstream performance of our models confirms that higher alignment leads to better performance and reduced hallucinations. Through our experiments, we also highlighted the need for a more comprehensive protein dataset that categorizes proteins based on their popularity and rarity, while also incorporating detailed and diverse descriptions on each protein. These insights pave the way for future research and development, helping to create more robust and accurate multimodal systems that can better understand and represent protein data.

\section*{Methods}

The development of protein-focused MLLMs has emerged as a promising direction for comprehensive protein analysis. These MLLMs aim to jointly process and understand both the structural characteristics and textual descriptions of proteins to enable more sophisticated analysis and prediction tasks. Two critical components in building effective protein MLLMs are GDMs, which capture the complex three-dimensional structural information of proteins, and LLMs, which process textual descriptions of protein properties and functions. The performance of these MLLMs heavily depends on how well the representations from GDMs and LLMs are aligned in a shared semantic space. However, current approaches to this alignment often rely on heuristic methods without a thorough understanding of the underlying factors that influence alignment quality.

Our methodology aims to systematically investigate this alignment challenge through three interconnected components: (1) data preprocessing to standardize different protein representations, (2) representation extraction that preserves both structural and semantic information, and (3) alignment techniques that bridge these different modalities while maintaining biological relevance. The overview of our methodology is shown in Figure \ref{fig:overview}. Through this approach, we seek to understand what factors contribute to successful alignment between GDM and LLM representations, how different protein properties affect alignment quality, and what strategies can improve alignment performance. This understanding is essential for developing more effective protein-focused MLLMs that can better integrate structural and textual information for advanced protein analysis tasks.

\vspace{3pt}
\noindent
\emph{Models:} 
In this paper, we selected four state-of-the-art GDMs specialized in the protein domain, as well as three pre-trained LLMs. Each model has a distinct embedding dimensionality, adding diversity to our experiments. For GDMs, we used the following models: GAT with an embedding size of 64 \cite{velickovic2018graph}, ScanNet with an embedding size of 128 \cite{tubiana2022scannet}, GVP with an embedding size of 148 \cite{jing2020learning}, and GearNet with an embedding size of 3072 \cite{zhangprotein}. All GDMs were pre-trained and loaded directly from their respective official repositories to ensure consistent and stable performance.
For LLMs, we selected Gemma2-2B with an embedding size of 2304 \cite{team2024gemma}, LLaMa3.1-8B with an embedding size of 4096 \cite{dubey2024llama}, and LLaMa3.1-70B with an embedding size of 8192. All LLMs were loaded directly from Huggingface's repository.

\vspace{3pt}
\noindent
\emph{Dataset:}
We downloaded protein data from the RCSB PDB database \cite{rose2016rcsb}. For this study, we randomly selected 20,000 protein samples. Each protein has two modalities: a text-based description from the FASTA file and a 3D structural graph modality from the PDB file. We randomly split the dataset into 80\% for training, 10\% for validation, and 10\% for testing. A more detailed statistic of the dataset is shown in Supplemental Note \ref{appendix:dataset}.

\subsection*{Step I: Data Preprocessing}
Our approach begins with the collection of protein data, denoted as \( {D} \). We ensure that each protein sample, \( d_i \in {D} \), contains both a FASTA file and a PDB file.

\subsubsection*{FASTA Preprocessing}
Since the selected LLMs are pre-trained on general text data without specific knowledge of proteins, we convert the FASTA file into a clear and detailed text description that the LLMs can understand. This is shown in Figure \ref{fig:overview} Data Preprocessing upper part. The process requires stripping unnecessary information and retaining only the essential content. A typical FASTA file contains several key details: protein ID, chains, molecule name, organism, and the amino acid sequence. While we consider most of this information important, the amino acid sequence itself is often lengthy and may introduce noise, so we exclude it from the text description and instead include the sequence length.
Additionally, the FASTA file allows us to determine whether a protein consists of a single chain or multiple chains. For single-chain proteins, we use regular expressions to extract relevant details. However, for multi-chain proteins, the complexity increases, and regular expressions alone often fail to capture all information. In such cases, we employ GPT-4o to generate accurate descriptions, extracting the chains and corresponding organisms. The GPT-4o prompt is provided in Supplemental Note \ref{appendix:gpt_prompt}.

After processing, each protein is described with text similar to: ``The protein structure [protein\_id] has a sequence length of [number] amino acids. Here is more information: The protein structure [protein\_id] involves the following chains: [chains]. The protein is named [protein\_name] and is derived from the organism [organism].'' We manually validated the GPT-generated descriptions for all multi-chain proteins and found them to be accurate compared to the original FASTA files. Examples of protein descriptions are shown in Supplemental Note \ref{appendix:protein_example}.

\subsubsection*{PDB Preprocessing}
PDB file preprocessing is more complex as it involves constructing appropriate graph structures for different GDM models. This is shown in Figure \ref{fig:overview} Data Preprocessing lower part.
For GearNet and ScanNet, both models accept PDB files directly and handle preprocessing internally, so no additional preprocessing is required on our end.
For the GAT model, the node feature and graph structure was parsed directly from the PDB file, which together form a graph \( G = (V, E) \). Each node \( i \in V \) represents an atom. The graph structure is defined by the edges \( (i, j) \in E \), which indicate the connections between the nodes.
For the GVP model, the protein PDB data is converted into a JSON format as shown below:
\begin{lstlisting}[language=json,firstnumber=1]
[{      `ID': `ID',
        `seq': `TQDCSFQHSP...',
        `coords': [[[74.46, 58.25, -21.65],...],...]}...]
\end{lstlisting}
Each protein includes three key elements: `ID' refers to the protein's id, `seq' represents the amino acid sequence, and `coords' stores the 3D coordinates of each residue in the protein's structure. Each coordinate is represented as a list of three floating-point values corresponding to the x, y, and z positions of the residue.

\subsection*{Step II: Latent Representation Extraction}
Once the data has been processed, the next step is to feed it into the respective models for latent representation extraction. This process is shown in Figure \ref{fig:overview} ``Representation Extraction'' part.

\subsubsection*{LLMs Representation}

For all three LLMs, we follow a consistent representation extraction procedure. First, we load the pre-trained models directly from Huggingface. The protein’s text description is tokenized using the model's associated tokenizer, converting it into input IDs that can be processed by the LLM. We then pass these tokenized inputs through the model and extract the hidden states from the final layer. Specifically, we select the representation of the last token, as it typically holds the most information from the protein’s description, given that it is derived from the final layer, which holds the richest contextual information. It is worth noting that for Gemma2-2B and LLaMa3.1-8B, we utilized the full-weight models. However, due to hardware constraints, LLaMa3.1-70B was loaded with float16 precision to fit within the available GPU memory.

\subsubsection*{GDMs Representation}
Different from LLMs, each GDM has its own way to extract representation. 

\begin{itemize}\setlength\itemsep{-0.3em}
\item{\emph{GearNet}:}
GearNet encodes protein structure information through a relational graph neural network (GNN) and an edge message-passing mechanism. The input to the model consists of a residue-level relational graph constructed from the 3D structure of the protein, which Gearnet's code will derived from the input PDB file automatically. The detail is shown in Supplemental Note \ref{appendix:gearnet}.
In the end, to obtain the final protein representation, the features from all hidden layers are concatenated, which results in the final protein representation has the shape \( [1, 3072] \).

\item{\emph{GVP}:}
Previously, we have processed the PDB file into the appropriate format supported by the GVP model, which includes a list of residues and their associated 3D coordinates. The detail of how we feed the data into GVP is shown in Supplemental Note \ref{appendix:gvp}.
In the end, the representation for each protein has the shape \( [1, \text{node\_size}, 148] \), where \( \text{node\_size} \) corresponds to the number of residues in the protein. Finally, to obtain a fixed-size representation for the entire protein, we apply average pooling across the node dimension. This results in a protein representation of shape \( [1, 148] \), which will be stored and used later.

\item{\emph{ScanNet}:}
To obtain the protein representation using ScanNet, we start by feeding the raw PDB file, which contains detailed information about the atomic coordinates of the protein. ScanNet will automatically preprocess and operates directly on the structure of the protein as described in the PDB file. The process can be divided into several key steps: parsing the PDB file, constructing atomic and amino acid representations, and passing these through ScanNet’s geometric deep learning architecture. The detail is shown in Supplemental Note \ref{appendix:scannet}.
In the end, the protein representation has the shape of \([1, \text{node\_size}, 128]\), where \(\text{node\_size}\) corresponds to the number of amino acids in the protein, and each amino acid has a 128-dimensional latent feature vector. Then, we use average pooling to reduce the \(\text{node\_size}\) dimension, resulting with \([1, 128]\) for each protein.

\item{\emph{GAT}:}
Previously, we already processed the protein PDB file into the format that GAT supports. Therefore, graph \( G \) is fed into a GAT model to compute latent representations. The detail is shown in Supplemental Note \ref{appendix:gat}.
In GAT architecture, they use 8 attention heads, with each head computing 8 features, leading to a concatenated output of size \( 64 \) for each node, which results in protein's final representation size of [1, node\_size, 64]. Then we simply perform average pooling to remove the `node\_size' dimension. This process yields the latent representation of the protein with size of [1, 64], which will be stored and used later.

\end{itemize}

\subsection*{Step III: Representation Alignment}

After we extracted protein representation from all models, we can combine these models into $12$ model pairs (i.e. one LLM and one GDM). For each model pairs, we train two projection heads, one for each model. All the projection head has the same structure, with one simple linear layer that map the model's embedding dimension (input dimension) to the LLM embedding dimension (output dimension). Then we normalize the output embedding. Therefore, after projection, the projected graph representation will have the same dimension as the projected text representation. This process is shown in Figure \ref{fig:overview} Representation Alignment.


Our objective is to maximize the cosine similarity to 1 between the projected graph representation and the projected text representation that belong to the same protein ID (positive pair). Conversely, we aim to keep the cosine similarity close to 0 for pairs where the projected graph and text representations originate from different protein IDs (negative pairs). It is important to note that we do not minimize the cosine similarity to -1, as a value of -1 would imply a strong negative correlation, which is not appropriate for unrelated protein alignment. A higher cosine similarity means a higher alignment score between two representations.

\subsubsection*{Contrastive Loss Function}

To maximize the cosine similarity of positive pairs to 1 and minimize that of negative pairs to 0, we apply a contrastive loss function during training. Specifically, we use a modified version of the InfoNCE loss \cite{oord2018representation}.
Let \( g_i \) be the projected graph representation for protein \( i \), \( t_j \) be the projected text representation for protein \( j \), and \( \text{sim}(g_i, t_j) \) represent the cosine similarity between \( g_i \) and \( t_j \), adjusted to the range \([0, 1]\) by using \( \frac{\text{sim}(g_i, t_j) + 1}{2} \). The temperature parameter \( \tau \) is set to 0.2 \cite{ngo2024language}, and the batch size \( B \) is 32.
For each protein \( i \), the positive pair similarity is given by \( \text{sim}^+_i = \frac{\text{sim}(g_i, t_i) + 1}{2} \), and the negative pair similarities are \( \text{sim}^-_{ij} = \frac{\text{sim}(g_i, t_j) + 1}{2} \) for all \( j \neq i \) within the batch.
After calculating these similarities, we apply exponentiation and scale them by the temperature \( \tau \):
\[
\mathcal{L}_{\text{total}} = \frac{1}{B} \sum_{i=1}^{B} -\log \left( \frac{\exp\left(\frac{\text{sim}(g_i, t_i) + 1}{2 \tau}\right)}{\exp\left(\frac{\text{sim}(g_i, t_i) + 1}{2 \tau}\right) + \sum_{j \neq i} \exp\left(\frac{\text{sim}(g_i, t_j) + 1}{2 \tau}\right)} \right).
\]
This overall loss function averages the loss over all proteins in the batch, effectively encouraging high similarity for positive pairs and low similarity for negative pairs. The detail of reweighting technique is shown in Supplemental Note \ref{appendix:reweighting}.

\subsubsection*{Implementation Details}

All projection heads were trained for 40 epochs. To ensure reproducibility, we set the random seed to 42. The learning rate was set to $1 \times 10^{-3}$, and the batch size was 32. We used Adam as the optimizer and applied model checkpointing to retaining the best projection heads (i.e. only store the model weight when the loss is decreased in validation set). We used four A100 GPUs and one A6000 GPU for hardware support. The A100 GPUs were primarily used for experiments involving the LLaMa3.1-70B model, while the A6000 GPU was sufficient for the remaining experiments.

\section*{Resource Availability}


\subsection*{Lead Contact}


Requests for further information and resources should be directed to and will be fulfilled by the lead contact, Dong Shu (dongshu2024@u.northwestern.edu).

\subsection*{Materials Availability}


This study did not generate new materials.

\subsection*{Data and Code Availability}


\begin{itemize}
    \item  Our data was downloaded directly from the RCSB PDB database \url{https://www.rcsb.org/} and are publicly available. All other data reported in this paper will be shared by the lead contact upon request.
    \item The code to reproduce the results of this article can be found at \url{https://github.com/Tizzzzy/LLM-GDM-alignment} and is publicly available. Additionally, we have archived the code on Zenodo \cite{tizzzzzzy_2025_14934851}, where it can be accessed via DOI: 10.5281/zenodo.14934851.
    
    \item Any additional information required to reanalyze the data reported in this paper is available from the lead contact upon request.    
\end{itemize}

\section*{Acknowledgments}
The work is in part supported by NSF \#2310261. The views and conclusions in this paper are those of the authors and should not be interpreted as representing funding agencies.



\section*{Author Contributions}



Conceptualization, D.S. and M.D.; Methodology, D.S. and M.D.; Software, D.S.; Investigation, D.S.; Resources, B.D. and M.D.; Writing – Original Draft, D.S., and M.D.; Writing – Review \& Editing, All authors; Supervision: K.G., K.Z., J.T., and M.D.

\section*{Declaration of Interests}


The authors declare no competing interests.

\section*{Declaration of Generative AI and AI-assisted technologies in the writing process}


The authors used ChatGPT in order to check grammar mistakes.

\section*{Supplemental Information}




\begin{description}
\item \textbf{1} Dataset Statistic
\item \textbf{2} Prompt Used in FASTA Preprocess
\item \textbf{3} Protein Examples
\item \textbf{4} GDM Representation Details
\item \textbf{5} Popular and Rare Protein
\item \textbf{6} GearNet Training
\item \textbf{7} Number of Projection Head Layer
\item \textbf{8} LLM Finetune Implementation Details
\item \textbf{9} Protein-focused MLLMs Methodology
\item \textbf{10} Reweighting Methodology
\item \textbf{11} Retrieval Methodology.

\item \textbf{Figure S1} Data Statistic.
\item \textbf{Figure S2} Single-Layer to Multi-Layer Projection Head: Architecture and Results.
\item \textbf{Figure S3} Architecture of Protein-Focused MLLMs and Their Generated Responses.
\item \textbf{Figure S4} Popular Protein Examples.
\item \textbf{Figure S5} Rare Protein Examples.
\item \textbf{Figure S6} Example of Protein-focused MLLMs Generated Output for Protein ID 3I1A.
\end{description}

\bibliography{references}

\newpage

\section*{Supplemental Notes}

\renewcommand{\thefigure}{S\arabic{figure}} 

\section{Dataset Statistic}
\label{appendix:dataset}

\begin{figure}[t]
    \centering
    \includegraphics[width=1\linewidth]{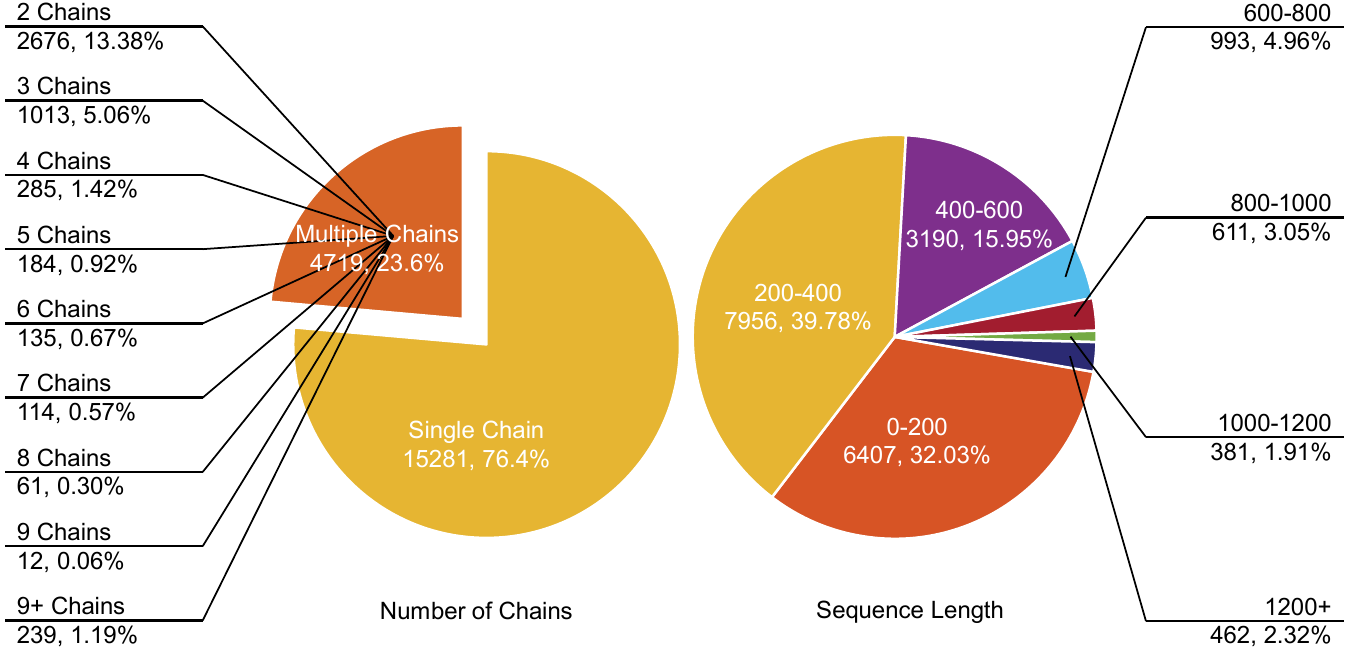}
    \caption{\textbf{Data Statistic.} The pie chart on the left categorizes proteins by the number of chains, while the pie chart on the right categorizes them by sequence length.}
    \label{fig:data_statistic}
\end{figure}

Our dataset consists of 20,000 proteins, each with an associated FASTA file and PDB file. Detailed statistics for the dataset are presented in Figure \ref{fig:data_statistic}, where each pie chart represents a different categorical breakdown of the same dataset. For the ``Number of Chains'' statistic, the chart shows the count of single-chain proteins versus multiple-chain proteins, with further subdivisions indicating the exact number of chains for multi-chain proteins. For the ``Sequence Length'' statistic, proteins are categorized by sequence length range. Each section of the pie chart is labeled with the relevant number of chains or sequence length range at the top, with the exact count and corresponding percentage displayed below each label.

\section{Prompt Used in FASTA Preprocess}
\label{appendix:gpt_prompt}

During the FASTA file preprocessing stage, we used GPT-4o to handle proteins with multiple chains. Below is the GPT prompt we employed for this task:

\begin{lstlisting}
==== System Prompt ==== 
You are a biologist with expertise in protein sequence analysis. 
Your task is to summarize complex protein sequence data into two or
three sentences that highlight key features such as molecute type, 
chains, structural motifs, organism, etc.

==== User Query ==== 
Summarize the following protein knowledge, start with the sentence: 
'The protein structure {protein_id} has a sequence length of: 
{sequence_length} amino acids.'
Here is more information about {protein_id}: 
{fasta_text}
\end{lstlisting}

\section{Protein Examples}
\label{appendix:protein_example}
As shown in Figure \ref{fig:popular_proteins} and \ref{fig:rare_proteins}, we listed several proteins and their corresponding description. The meaning of `popular' and `rare' protein is discussed in Research Question 3: Protein Perspective Analysis.

\section{GDM Representation Details}

\subsection{GearNet}
\label{appendix:gearnet}
\paragraph{Protein Graph Construction:}
The structure of a protein is represented as a residue-level relational graph \( G = (V, E, R) \), where \( V \) represents the set of nodes, \( E \) the set of edges, and \( R \) the set of edge types. Each node \( v_i \in V \) corresponds to a residue in the protein, with its 3D coordinates \( \mathbf{x}_i \in \mathbb{R}^3 \), and each edge \( e_{ij} \in E \) represents a relationship between residues based on either sequential proximity or spatial proximity. In particular, GearNet incorporates three types of edges:
\begin{itemize}
    \item \textbf{Sequential edges:} connect residues within a distance of 2 in the sequence.
    \item \textbf{Radius edges:} connect residues whose \( C_\alpha \) atoms are within a given radius in 3D space.
    \item \textbf{K-nearest neighbor edges:} connect each residue to its \( k \)-nearest neighbors in 3D space.
\end{itemize}
\noindent
Each node \( v_i \) is initially represented by its residue type and spatial coordinates, while each edge \( e_{ij} \) is represented by its edge type (sequential or spatial) and spatial distance.

\paragraph{Relational Graph Convolution:}
GearNet applies relational message passing on the constructed protein graph, leveraging both node and edge features. The relational graph convolutional layer is defined as:
\[
h^{(0)}_i = f_i, \quad u^{(l)}_i = \sigma\left( \text{BN} \left( \sum_{r \in R} W_r \sum_{j \in \mathcal{N}_r(i)} h^{(l-1)}_j \right) \right), \quad h^{(l)}_i = h^{(l-1)}_i + u^{(l)}_i
\]
where \( f_i \) is the initial feature of node \( i \), \( \mathcal{N}_r(i) \) is the set of neighbors connected to node \( i \) via edges of type \( r \), \( W_r \) is a learnable weight matrix for edge type \( r \), and \( \sigma \) is a ReLU activation function. Batch normalization (BN) is applied for normalization across batches, and a residual connection is added from the previous layer to stabilize training. This relational message passing allows GearNet to effectively capture both sequential and spatial information in the protein structure, as different edge types are modeled with different convolutional kernels.

\paragraph{Edge Message Passing:}
In addition to node message passing, GearNet incorporates an edge message-passing mechanism to explicitly model interactions between edges. The model constructs an edge-level graph \( G' = (V', E', R') \), where each node in \( G' \) corresponds to an edge in the original protein graph \( G \). The edge message passing is defined as:
\[
m^{(0)}_{(i,j,r)} = f_{(i,j,r)}, \quad m^{(l)}_{(i,j,r)} = \sigma\left( \text{BN} \left( \sum_{r' \in R'} W'_{r'} \sum_{(w,k,r') \in \mathcal{N}'_{r'}((i,j,r))} m^{(l-1)}_{(w,k,r')} \right) \right)
\]
where \( m^{(l)}_{(i,j,r)} \) represents the message for edge \( (i,j,r) \) at layer \( l \), \( \mathcal{N}'_{r'}((i,j,r)) \) is the set of neighboring edges in \( G' \), and \( W'_{r'} \) is a learnable weight matrix for edge type \( r' \). The angular information between edges is used to determine edge types, with edges having smaller angles expected to interact more strongly. The output from the edge message passing is integrated into the node update step by modifying the aggregation function as follows:
\[
u^{(l)}_i = \sigma\left( \text{BN} \left( \sum_{r \in R} W_r \sum_{j \in \mathcal{N}_r(i)} \left( h^{(l-1)}_j + \text{FC}(m^{(l)}_{(j,i,r)}) \right) \right) \right)
\]
where \( \text{FC} \) is a fully connected layer applied to the edge message.

\paragraph{Protein Representation:}
After multiple rounds of relational graph convolution and edge message passing, each residue \( v_i \) in the protein is represented by a feature vector \( h^{(L)}_i \), where \( L \) is the number of layers in the model. In the original GearNet model, the hidden dimensions are set to \( \text{hidden\_dims} = [512, 512, 512, 512, 512, 512] \), meaning that each residue is represented by a feature vector of size 512 after each layer. To obtain the final protein representation, the features from all layers are concatenated:
\[
h_{\text{protein}} = \left[ h^{(1)}_i, h^{(2)}_i, \dots, h^{(6)}_i \right] \in \mathbb{R}^{3072}
\]
where the concatenation of six layers (each of dimension 512) results in a 3072-dimensional feature vector for each protein. The final protein representation has the shape \( [1, 3072] \).

\subsection{GVP}
\label{appendix:gvp}
\paragraph{Graph Construction:}
Using the JSON input format, each protein's backbone structure is parsed to identify each residue and its corresponding coordinates for the N, C-alpha, C, and O atoms. This [num\_residues x 4 x 3] nested list provides precise spatial positioning essential for structural insights. For each residue, a node \( v_i \in V \) is created in the graph \( G \), with scalar and vector features, and edges \( e_{ij} \in E \) are created between nodes based on spatial proximity between residues. Scalar features \( h_s \in \mathbb{R}^{n_s} \) include properties such as the amino acid type and backbone dihedral angles (e.g., \( \phi, \psi, \omega \)). Vector features \( h_v \in \mathbb{R}^{n_v \times 3} \) encode geometric information, such as the direction of the residue's neighbors. These features are processed through a series of GVP layers, which operate on both scalar and vector features to propagate information through the graph.

\paragraph{Message Passing and Geometric Vector Perceptrons:}
The core of the GVP model is its message-passing architecture, which updates both scalar and vector features at each graph propagation step. For each node \( v_i \), the message from a neighboring node \( v_j \) is computed using the scalar and vector features of both nodes and their connecting edge:
\[
h_{m}^{(j \to i)} = \text{GVP}\left(\left[h_s^{(j)}, h_v^{(j)}, h_{e}^{(j \to i)}\right]\right)
\]
where \( h_s^{(j)} \) and \( h_v^{(j)} \) are the scalar and vector features of node \( j \), and \( h_{e}^{(j \to i)} \) is the edge embedding, which includes both distance and direction information. The GVP function processes these features, maintaining rotation-equivariance for the vector features:
\[
h_{v'}^{(i)} = \sigma\left(\sum_{j \in \mathcal{N}(i)} \mathbf{W}_v h_v^{(j)}\right), \quad h_{s'}^{(i)} = \sigma\left(\mathbf{W}_s h_s^{(i)} + \|\mathbf{W}_v h_v^{(j)}\|\right)
\]
where \( \mathbf{W}_v \) and \( \mathbf{W}_s \) are learnable weight matrices for vector and scalar features, respectively, and \( \sigma \) is a non-linear activation function.

\paragraph{Protein Representation:}
After multiple rounds of message passing, each node \( v_i \) is associated with a scalar feature vector \( h_s^{(i)} \in \mathbb{R}^{100} \) and a vector feature matrix \( h_v^{(i)} \in \mathbb{R}^{16 \times 3} \). To construct the final protein representation, the scalar and vector features are combined. The vector features are reduced to scalars by taking their L2-norm:
\[
h_v^{(i)} = \|\mathbf{h_v^{(i)}}\|_2
\]
The final per-node embedding is then formed by concatenating the scalar and vector features:
\[
h^{(i)} = \left[ h_s^{(i)}, h_v^{(i)} \right] \in \mathbb{R}^{148}
\]
where \( 148 \) is the dimension of the combined scalar and vector features (100 scalar channels and 3 vector channels of size 16, i.e., \( 3 \times 16 + 100 = 148 \)). Thus, the representation for the entire protein has the shape \( [1, \text{node\_size}, 148] \), where \( [\text{node\_size}] \) corresponds to the number of residues in the protein. Finally, to obtain a fixed-size representation for the entire protein, we apply average pooling across the node dimension:
\[
h_{\text{protein}} = \frac{1}{\text{node\_size}} \sum_{i=1}^{\text{node\_size}} h^{(i)}
\]
This results in a protein representation of shape \( [1, 148] \), which will be stored and used later.

\subsection{ScanNet}
\label{appendix:scannet}

\paragraph{Parsing the PDB File:}
First, the PDB file is parsed to extract the protein's amino acid sequence and atomic point cloud. Each atom is represented as a triplet:
\[
\{ (\mathbf{x}_l, \text{id}_{\text{residue}_l}, \text{id}_{\text{atom}_l}) \mid l \in [1, N_{\text{atoms}}] \}
\]
where \( \mathbf{x}_l \in \mathbb{R}^3 \) represents the atomic coordinates, and \( \text{id}_{\text{residue}_l} \) and \( \text{id}_{\text{atom}_l} \) represent the residue and atom IDs, respectively. Only heavy atoms belonging to classical residues are considered.

\paragraph{Local Frame Construction:} 
Next, a local reference frame is defined for each heavy atom based on the molecular graph of the protein (i.e., nodes represent atoms and edges represent covalent bonds). For each atom, we select two neighboring atoms to construct a local coordinate system. The center is defined by the atom itself, and the orientation is determined by its neighbors using the following equations:
\[
f_{l1} = \mathbf{x}_{\text{atom}_l}
\]
\[
f_{l4} = \frac{\mathbf{x}_{\text{neighbor}_3} - \mathbf{x}_{\text{atom}_l}}{\|\mathbf{x}_{\text{neighbor}_3} - \mathbf{x}_{\text{atom}_l}\|}
\]
\[
f_{l3} = \frac{f_{l4} \times (\mathbf{x}_{\text{neighbor}_2} - \mathbf{x}_{\text{atom}_l})}{\| f_{l4} \times (\mathbf{x}_{\text{neighbor}_2} - \mathbf{x}_{\text{atom}_l})\|}
\]
\[
f_{l2} = \frac{f_{l3} \times f_{l4}}{\|f_{l3} \times f_{l4}\|}
\]
This frame is then used to transform atomic neighborhoods into a consistent local coordinate system.

\paragraph{Atomic and Amino Acid Pooling:} 
ScanNet constructs atomic representations by extracting the local neighborhood of each atom and encoding it with spatio-chemical filters. These filters operate on the atomic coordinates and attributes (such as atom type) and produce an atomic-scale representation. This process is mathematically formalized using spatio-chemical Gaussian filters:
\[
y_m = \text{ReLU} \left( \sum_{k,g,n} W_{mgn} G(\mu_g, \Sigma_g, x_k) a_k \right)
\]
where \(G(\mu_g, \Sigma_g, x_k)\) is the Gaussian kernel parameterized by mean \( \mu_g \) and covariance \( \Sigma_g \), applied to the local coordinates \( x_k \) of the neighborhood. The atomic-scale representations are then pooled at the amino acid level. The pooling step aggregates the atom-wise features into a single representation for each amino acid using a multi-head attention mechanism:
\[
\mathbf{h}_{\text{AA}} = \sum_{i=1}^{N_{\text{atoms}}} \alpha_i \mathbf{h}_{\text{atom}_i}
\]
where \( \alpha_i \) are learned attention weights that determine the contribution of each atom to the amino acid representation.

\paragraph{Amino Acid Neighborhood Embedding and Protein Representation:} 
For each amino acid, its local neighborhood is constructed based on the \( C_\alpha \) atom, sidechain orientation, and backbone orientation. The same spatio-chemical filtering process is applied to obtain an amino acid-wise representation:
\[
\mathbf{h}_{\text{AA}} = \text{ReLU} \left( \sum_{k,g,n} W_{mgn} G(\mu_g, \Sigma_g, x_k) a_k \right)
\]
where \( G(\mu_g, \Sigma_g, x_k) \) is the Gaussian kernel applied to the neighborhood of each amino acid. The amino acid representations are then aggregated to form a protein-level representation. This representation has the shape of \([1, \text{node\_size}, 128]\), where \(\text{node\_size}\) corresponds to the number of amino acids in the protein, and each amino acid has a 128-dimensional latent feature vector. Then, we simply use the average pooling to reduce the \(\text{node\_size}\) dimension, leaving us with \([1, 128]\) for each protein:
\[
\mathbf{h}_{\text{protein}} = \frac{1}{\text{node\_size}} \sum_{i=1}^{\text{node\_size}} \mathbf{h}_{\text{AA}_i}
\]

\subsection{GAT}
\label{appendix:gat}

For a graph with \( N \) nodes, the input to the GAT layer is a set of node features \( \mathbf{h} = \{\mathbf{h}_1, \mathbf{h}_2, \dots, \mathbf{h}_N\} \), where \( \mathbf{h}_i \in \mathbb{R}^F \) is the feature vector of node \( i \), and \( F \) is the number of features associated with each atom. Each node is first transformed linearly using a shared learnable weight matrix \( W \in \mathbb{R}^{F' \times F} \), yielding: \(\mathbf{h}_i' = W \mathbf{h}_i\). Then, the attention mechanism computes attention coefficients for each pair of nodes \( i \) and \( j \), where \( j \in \mathcal{N}_i \) is a neighboring node of \( i \). The attention coefficients \( e_{ij} \) are given by:
$
e_{ij} = \text{LeakyReLU}\left(\mathbf{a}^T \left[W \mathbf{h}_i \parallel W \mathbf{h}_j\right]\right)
$,
where \( \mathbf{a} \in \mathbb{R}^{2F'} \) is a learnable weight vector, and \( \parallel \) denotes concatenation.
These attention coefficients are then normalized across all neighbors of node \( i \) using the softmax function:
$
\alpha_{ij} = \frac{\exp(e_{ij})}{\sum_{k \in \mathcal{N}_i} \exp(e_{ik})}.
$
The normalized attention coefficients are used to compute the updated feature representation of node \( i \) as a weighted sum of its neighbors' features:
$
\mathbf{h}_i'' = \sigma\left(\sum_{j \in \mathcal{N}_i} \alpha_{ij} W \mathbf{h}_j\right)
$
where \( \sigma \) is a non-linear activation function.

To stabilize the learning process, GAT employs multi-head attention, where \( K \) independent attention mechanisms compute the attention coefficients and aggregate the features. The outputs from the different attention heads are concatenated as:
$
\mathbf{h}_i^{\text{multi}} = \big\|_{k=1}^K \sigma\left(\sum_{j \in \mathcal{N}_i} \alpha_{ij}^k W^k \mathbf{h}_j\right).
$
For the final prediction layer, the attention heads are averaged rather than concatenated:
$
\mathbf{h}_i^{\text{final}} = \sigma\left(\frac{1}{K} \sum_{k=1}^K \sum_{j \in \mathcal{N}_i} \alpha_{ij}^k W^k \mathbf{h}_j\right).
$
In GAT architecture, they use \( K = 8 \) attention heads, with each head computing \( F' = 8 \) features, leading to a concatenated output of size \( 64 \) for each node, which results in protein's final representation size of [1, node\_size, 64]. Then we simply perform average pooling to remove the `node\_size' dimension. This process yields the latent representation of the protein with size of [1, 64], which will be stored and used later.

\section{Popular and Rare Protein}
\label{appendix:protein}
\begin{figure}[hb]
    \centering
    \includegraphics[width=0.7\linewidth]{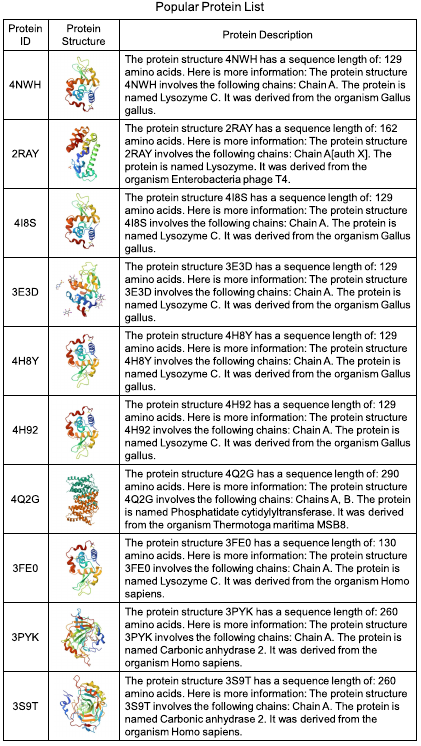}
    \caption{\textbf{Popular Protein Examples.} The figure provides details of 10 popular proteins. The first column lists the protein IDs, the second column shows the 3D structures of the proteins, and the third column contains the text descriptions for each protein.}
    \label{fig:popular_proteins}
\end{figure}

\begin{figure}[hb]
    \centering
    \includegraphics[width=0.8\linewidth]{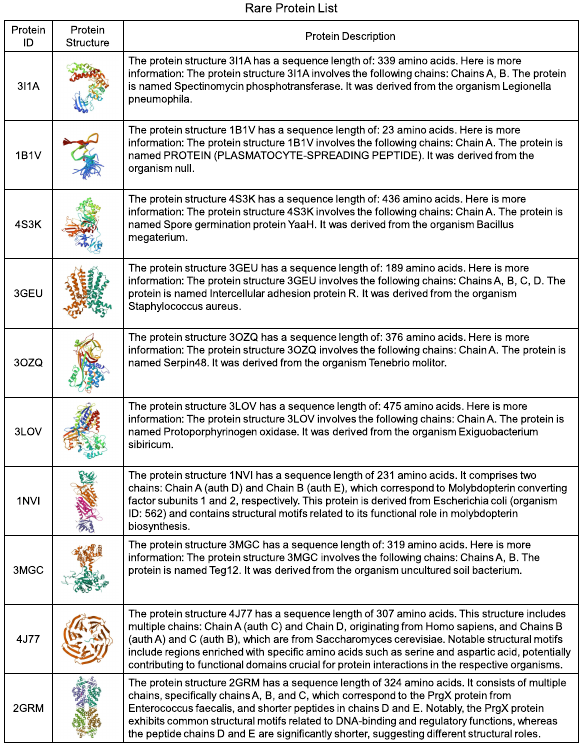}
    \caption{\textbf{Rare Protein Examples.} The figure provides details of 10 rare proteins. The first column lists the protein IDs, the second column shows the 3D structures of the proteins, and the third column contains the text descriptions for each protein.}
    \label{fig:rare_proteins}
\end{figure}

For the protein perspective analysis, we randomly selected 10 popular proteins and 10 rare proteins to study. The IDs of the popular proteins are: [4NWH, 2RAY, 4I8S, 3E3D, 4H8Y, 4H92, 1B1V, 3FE0, 3PYK, 3S9T], while the IDs of the rare proteins are: [3I1A, 4Q2G, 4S3K, 3GEU, 3OZQ, 3LOV, 1NVI, 3MGC, 4J77, 2GRM]. Details for these proteins are presented in Figure \ref{fig:popular_proteins} and \ref{fig:rare_proteins}.

Before random selection, we first collected all proteins' molecule names and organisms. We then used regular expressions to clean the names, excluding specific terms like numbers within parentheses, and combined the molecule names and organisms into distinct category labels. Next, we mapped all the proteins to these categories and calculated the count for each. Proteins within categories with higher counts were considered as popular, while those in categories with lower counts (or a count of one) were considered rare. We then organized the proteins in ascending order from rarest to most popular and randomly selected 10 popular proteins from the top 100 most popular categories and 10 rare proteins from the top 100 rarest categories.

\section{GearNet Training}
\label{appendix:gearnet_train}
To analyze whether the representation dimension size of the Geometric Deep Model (GDM) affects the alignment performance of the model pair, we trained GearNet using the Enzyme Commission dataset, following the tutorial steps from TorchDrug. We followed the original GearNet model configuration with an input dimension of 21, 7 node relations, 59 edge features, and 8 angle bins. The hidden layer size was adjusted to control the representation dimension, using the following configurations: \{[64], [128], [256], [512], [512, 512], [512, 512, 512, 512, 512, 512]\}. This resulted in six different dimension sizes: 64, 128, 256, 512, 1024, and 3072. We stopped at 3072, as it matches the dimension size of the original pretrained GearNet model. Batch normalization and a shortcut mechanism were applied, along with a sum-based readout layer for node feature aggregation. The model was trained using a binary cross-entropy loss function and optimized with the Adam optimizer. We evaluated performance using the AUPRC@micro and F1\_max metrics. The training was conducted for 10 epochs with a batch size of 4.

\section{Number of Projection Head Layer}
\label{appendix:proj_head_layer}
\begin{figure}
    \centering
    \includegraphics[width=1\linewidth]{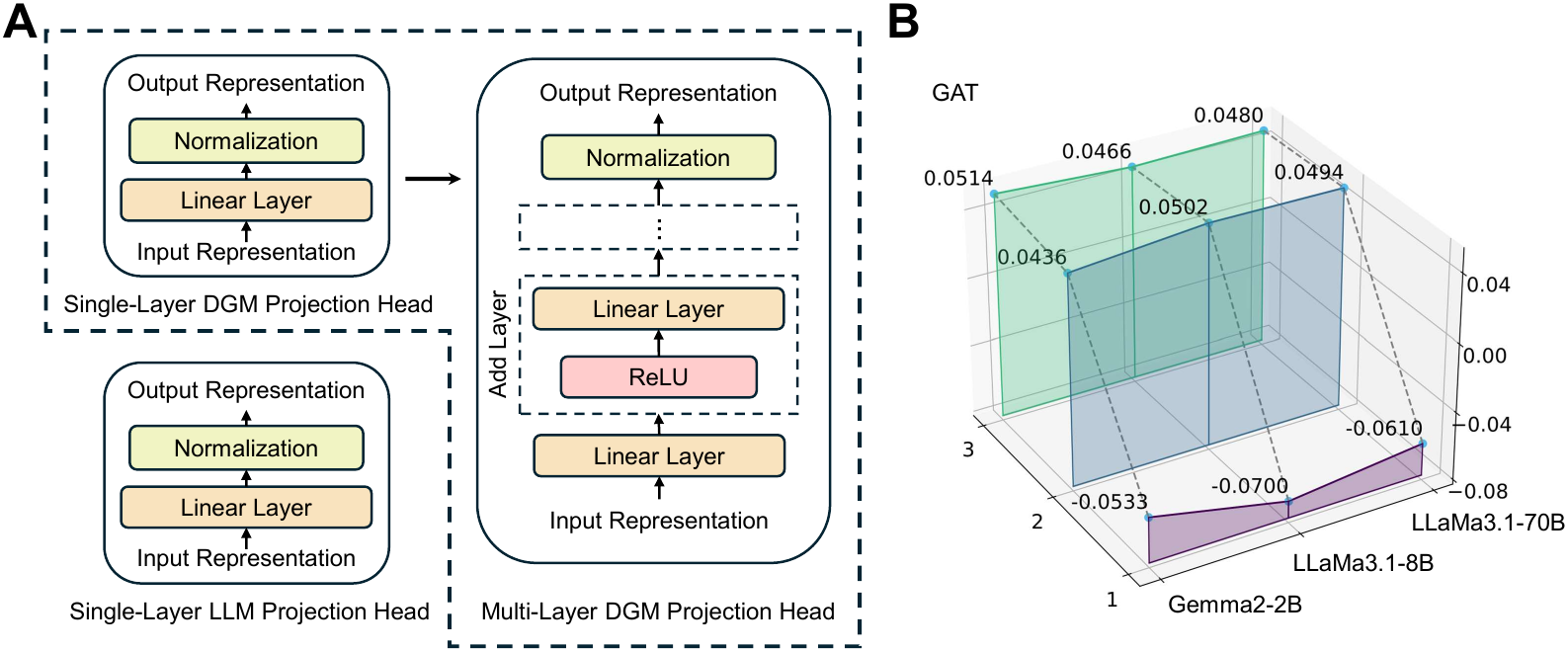}
    \caption{\textbf{Single-Layer to Multi-Layer Projection Head: Architecture and Results.} (A) The figure illustrates the shift from a single-layer to a multi-layer projection head architecture. Note, layers are only added to the GDM projection head, while the LLM projection head remains unchanged. (B) This extends the example from Figure \ref{fig:protein_combine_2}A, showing the alignment performance when additional layers are added to the projection head for model pairs involving GAT.}
    \label{fig:proj_head_detail}
\end{figure}

To address Research Question 5, we investigate whether increasing the number of layers in the projection head improves alignment performance. As shown in Figure \ref{fig:proj_head_detail}A, we only modify the number of linear layers in the GDM’s projection head, while leaving the LLM’s projection head unchanged. Our multi-layer projection heads consist of either 2 or 3 linear layers, with ReLU activation functions between each layer. Since each GDM has a different representation dimension, and the output dimension depends on the LLM used in the model pair, we selected different hidden layer dimensions for each GDM's projection head to ensure appropriate alignment.

For model pairs using Gemma2-2B as the LLM, the output dimension for all GDMs is set to match the Gemma2-2B embedding size, which is [2304]. In the 2-layer projection head for GearNet (original dimension [3072]), we chose [2560] as the hidden layer dimension, calculated as \(3072 - 512 = 2560\), to approximate the midpoint between [3072] and [2304]. For other GDMs with smaller original dimensions, we used a hidden dimension of [1024]. In the 3-layer projection head for GearNet, the hidden layer dimensions were set to [2816, 2560], while for other GDMs, we used [512, 1024].

For model pairs using LLaMa3.1-8B as the LLM, the output dimension was set to [4096] to match the LLM’s embedding size. In the 2-layer projection head, GearNet used [3584] as the hidden dimension, calculated as \(3072 + 512 = 3584\). For other GDMs, we used a hidden dimension of [2048]. In the 3-layer projection head, the hidden dimensions for GearNet were set to [3584, 3840], while for other GDMs, we used [512, 2048].

For model pairs using LLaMa3.1-70B as the LLM, the output dimension was set to [8192]. In the 2-layer projection head, we used a hidden dimension of [4096] for all GDMs. In the 3-layer projection head, GearNet used hidden dimensions of [4096, 6144], and other GDMs used [1024, 4096].

\section{LLM Finetune Implementation Details}
\label{appendix:llm_finetune}
To answer Research Question 6, we fine-tune all three of the LLMs, Gemma2-2B, LLaMa3.1-8B, LLaMa3.1-70B on the protein dataset we constructed during the FASTA preprocessing using LoRA (Low-Rank Adaptation) \cite{hu2021lora} to efficiently adjust the model's parameters.

First, the LLM and its corresponding tokenizer were loaded using HuggingFace. The model was loaded with 4-bit quantization to enable efficient memory usage while preserving the model's performance. Quantization techniques, such as nf4 (normal float 4-bit) and float16 computation, were employed to reduce the computational load during training. The model’s configurations were adjusted to disable caching and to use a single pretraining tensor parallelism thread. The input data for fine-tuning was formated like this:
\begin{lstlisting}
<|im_start|>user
What is the protein: {protein_id}?
<|im_end|>

<|im_start|>assistant
{protein_description}
<|im_end|>
\end{lstlisting}

For the fine-tuning process, LoRA was applied to enable efficient parameter adjustment without the need to fully fine-tune the LLM. The LoRA configuration specified a low-rank parameter r=8, with lora\_alpha=16 and a dropout of 0.05, targeting causal language modeling tasks. The training arguments were set to use the paged AdamW optimizer in 32-bit mode, with a learning rate of 2e-4 and a cosine learning rate scheduler. The training ran for 10 epoch with a batch size of 1 per device and gradient accumulation steps of 16 to account for memory constraints. Mixed precision (FP16) was used to further optimize memory usage and speed. We used four A100 GPUs and one A6000 GPU for hardware support. The A100 GPUs were used to fine-tune LLaMa3.1-70B, while the A6000 GPU was sufficient for the remaining experiments.

\section{Protein-focused MLLMs Methodology}
\label{appendix:protein_llm}

\begin{figure}
    \centering
    \includegraphics[width=0.75\linewidth]{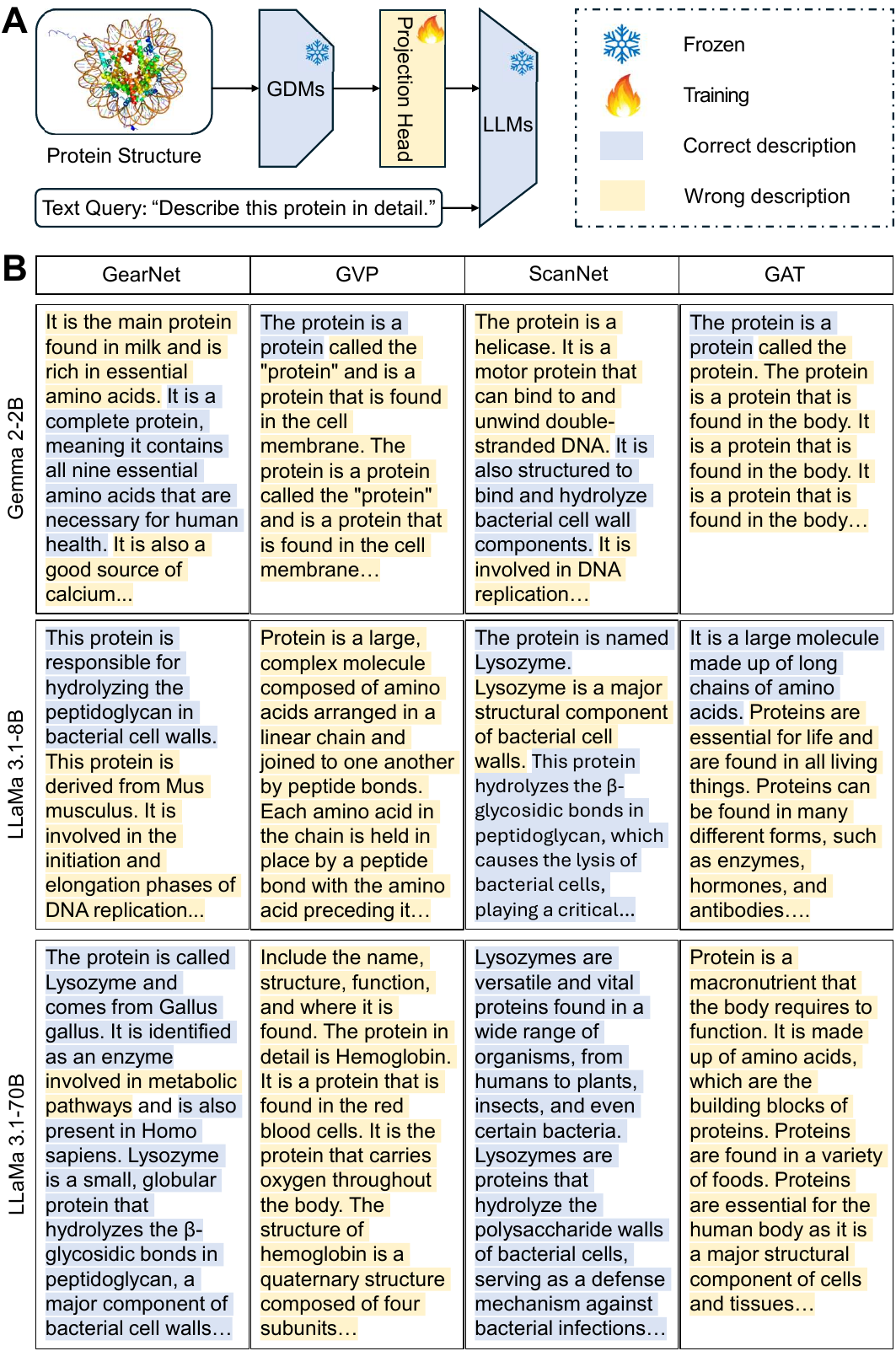}
    \caption{\textbf{Architecture of Protein-Focused MLLMs and Their Generated Responses.} (A) The architecture of a protein-focused MLLM comprises three primary components: a Geometric Deep Model (GDM), a projection head, and a Large Language Model (LLM). The GDM processes the 3D protein structures as its input. Subsequently, the projection head transforms these structural representations into the same dimension as the LLM. Then LLM takes both projected representation and textual inputs to generate its outputs. (B) The figure shows protein-focused MLLMs example responses for protein ID 4NWH. In the output, factually correct text is highlighted in blue, whereas text that is incorrect or involves hallucination is marked in yellow.}
    \label{fig:protein-LLM}
\end{figure}

As shown in Figure \ref{fig:protein-LLM}, our protein-focused MLLM contains three key components: a GDM, a trained projection head, and a LLM. The protein-focused MLLM can take two inputs: a protein structure and a text question. In this study, we focus on the protein description task, where the text input is ``Describe the protein in detail.'' First, the protein structure is processed by the GDM to extract a latent representation. This representation is fed into the trained projection head, which maps it to the same dimensional space as the LLM’s embeddings. The text input is tokenized and passed through the LLM to extract its representation. We then concatenate the projected protein representation with the text representation and feed them into the LLM to generate the output. This output will be used when we further fine-tuned the projection head to improve alignment between the protein and text representations. We use the protein descriptions that generated during the FASTA preprocessing section as the ground truth. We fine-tuned the projection head using Cross-Entropy Loss, which calculates the discrepancy between the predicted token sequence and the ground truth sequence:
\[
\mathcal{L} = - \frac{1}{N} \sum_{i=1}^{N} \sum_{t=1}^{T} \log P\left(y_t^{(i)} \mid x^{(i)}, y_{<t}^{(i)}\right)
\]
where \(N\) is the number of samples in the batch. \(T\) is the length of the ground truth description for the \(i\)-th sample. \(y_t^{(i)}\) is the ground truth token at position \(t\) for the \(i\)-th sample. \(x^{(i)}\) is the input to the model. \(y_{<t}^{(i)}\) is all ground truth tokens generated prior to \(t\) for the \(i\)-th sample.
During fine-tuning, the GDM and LLM parameters were frozen, and only the projection head was updated. We used the same implementation details as in the Representation Alignment section. The projection head was fine-tuned for 40 epochs with learning rate of $1 \times 10^{-3}$ and batch size of 32.

\section{Reweighting Methodology}
\label{appendix:reweighting}

We reweight the penalty of the rare protein in the loss function when training the projection head. The process starts by labeling protein as rare and popular. We used the same way as discussed in Supplemental Note \ref{appendix:protein} to rank top 100 rare protein from the training dataset. During the training, we reweight the penalty of these 100 rare protein in the loss function, by multiplying rare protein penalties by 2, and leave other protein penalties unchanged. If protein $i$ in 100 rare protein: 
\[
\mathcal{L}_{\text{total}} = \frac{1}{B} \sum_{i=1}^{B} \left(-\log \left( \frac{\exp\left(\frac{\text{sim}(g_i, t_i) + 1}{2 \tau}\right)}{\exp\left(\frac{\text{sim}(g_i, t_i) + 1}{2 \tau}\right) + \sum_{j \neq i} \exp\left(\frac{\text{sim}(g_i, t_j) + 1}{2 \tau}\right)} \right) * 2\right)
\]

\section{Retrieval Methodology}
\label{appendix:retrieval}

The retrieval process begins by mapping all training protein representations to the same dimensional space as the LLM using our reweighting-trained projection head. This mapped representation space is denoted as \( R \). Once all protein representations in the training dataset are mapped, each protein \( j \) in the testing set is also mapped into \( R \). For each protein \( j \), we identify its top \( k \) most similar proteins in the representation space \( R \) based on the cosine similarity with other protein \( i \):
\[
\text{Retrieved\_proteins}(j) = \text{Top}_k \big(\text{sim}(i, j) \mid i \in R\big).
\]
In this study, we evaluate \( k \) with different values: \([3, 5, 10]\). For each test protein \( j \), the ground truth descriptions of its top \( k \) similar proteins are prepended to the original input and fed into the protein-focused MLLM. The model's performance is then evaluated using the ROUGE and BLEU metrics.

\begin{figure}
    \centering
    \includegraphics[width=0.75\linewidth]{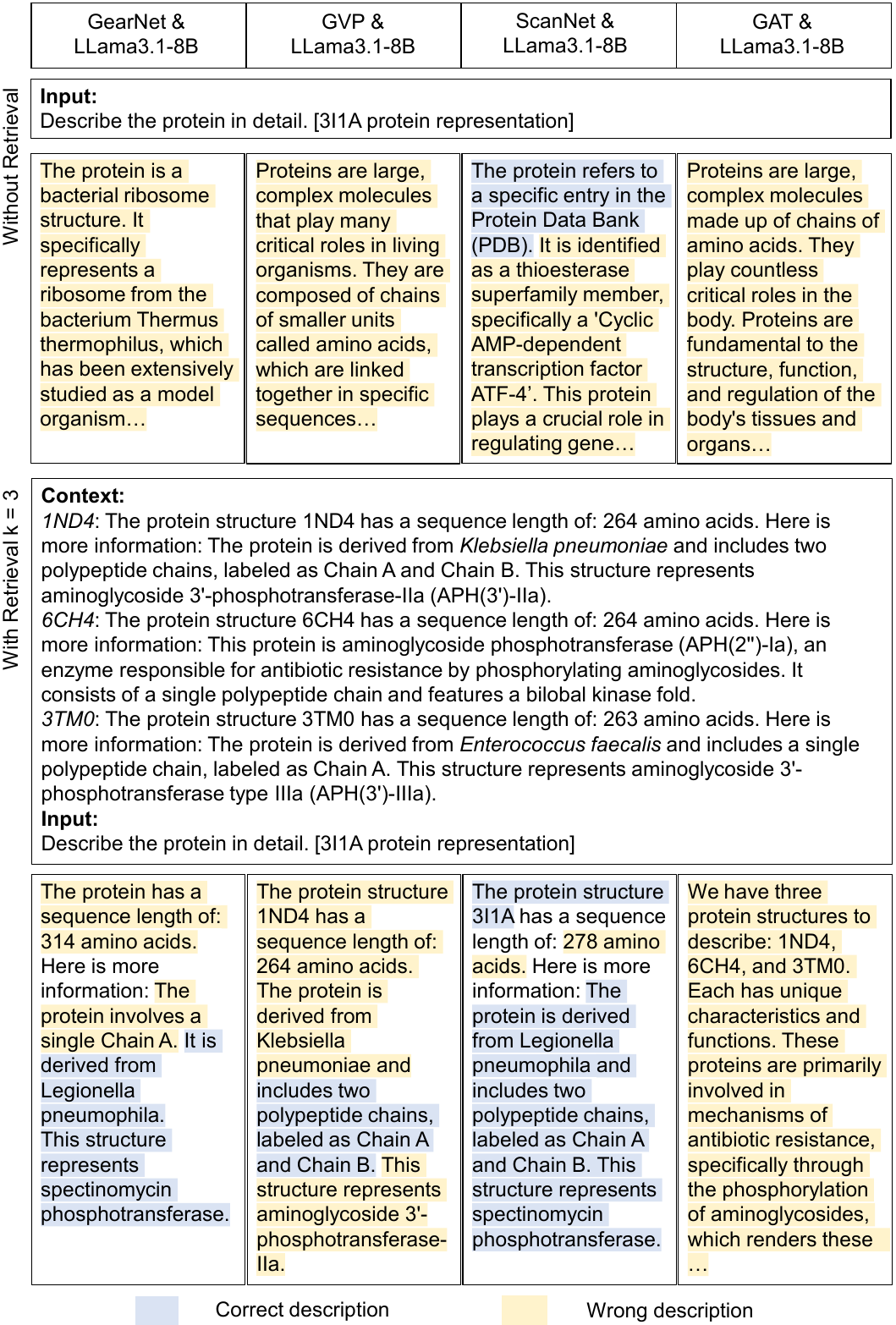}
    \caption{\textbf{Example of Protein-focused MLLMs Generated Output for Protein ID 3I1A.} The upper figure illustrates the output generated by the model using only the reweighting technique. The lower figure, on the other hand, displays the output generated by the model when both reweighting and retrieval techniques are employed.}
    \label{fig:reweight_retrieval}
\end{figure}

\end{document}